\pdfoutput=1
\documentclass[sigconf]{acmart}

\AtBeginDocument{%
  }

\definecolor{c1}{HTML}{1F9856}
\definecolor{c2}{HTML}{BE0D16}

\usepackage{bm}
\usepackage{ulem}
\usepackage{cases}
\usepackage{multirow}
\usepackage{nicematrix} %

\copyrightyear{2026}
\acmYear{2026}
\setcopyright{cc}
\setcctype{by}
\acmConference[KDD '26]{Proceedings of the 32nd ACM SIGKDD Conference on Knowledge Discovery and Data Mining V.1}{August 09--13, 2026}{Jeju Island, Republic of Korea}
\acmBooktitle{Proceedings of the 32nd ACM SIGKDD Conference on Knowledge Discovery and Data Mining V.1 (KDD '26), August 09--13, 2026, Jeju Island, Republic of Korea}
\acmPrice{}
\acmDOI{10.1145/3770854.3783959}
\acmISBN{979-8-4007-2258-5/2026/08}

\acmSubmissionID{123-A56-BU3}

\begin{document}

\title[FilDeep: Learning Large Deformations of Elastic-Plastic Solids with Multi-Fidelity Data]{FilDeep: Learning Large Deformations of Elastic-Plastic Solids with Multi-Fidelity Data}

\author{Jianheng Tang}
\orcid{0000-0002-4762-5943}
\affiliation{%
  \institution{Peking University}
  \department{School of Computer Science}
  \city{Beijing}
  \country{China}
}
\email{tangentheng@gmail.com}

\author{Shilong Tao}
\orcid{0009-0003-2497-3413}
\affiliation{%
  \institution{Peking University}
  \department{School of Computer Science}
  \city{Beijing}
  \country{China}
}
\email{shilongtao@stu.pku.edu.cn}

\author{Zhe Feng}
\orcid{0009-0000-8682-3769}
\affiliation{%
  \institution{Peking University}
  \department{School of Computer Science}
  \city{Beijing}
  \country{China}
}
\email{zhe.feng27@stu.pku.edu.cn}

\author{Haonan Sun}
\orcid{0009-0007-8784-6963}
\affiliation{%
  \institution{Peking University}
  \department{School of Computer Science}
  \city{Beijing}
  \country{China}
}
\email{sunhaonan331@pku.edu.cn}

\author{Menglu Wang}
\orcid{0009-0002-1306-3169}
\affiliation{%
  \institution{Fuyao Group}
  \city{Fuqing}
  \country{China}
}
\email{menglu.wang@fuyaogroup.com}

\author{Zhanxing Zhu}
\orcid{0000-0002-2141-6553}
\affiliation{%
  \institution{University of Southampton}
  \department{School of Electrical and Computer Science}
  \city{Southampton}
  \country{United Kingdom}
}
\email{z.zhu@soton.ac.uk}

\author{Yunhuai Liu}
\authornote{Corresponding author.}
\orcid{0000-0002-1180-8078}
\affiliation{
  \institution{Peking University}
  \department{School of Computer Science}
  \city{Beijing}
  \country{China}
}
\affiliation{%
  \institution{Key Lab of High Confidence Software Technologies (Peking University) \\ Ministry of Education}
  \city{Beijing}
  \country{China}
}
\email{yunhuai.liu@pku.edu.cn}

\renewcommand{\shortauthors}{Jianheng Tang et al.}

\begin{abstract}
The scientific computation of large deformations in elastic-plastic solids is crucial in various manufacturing applications.
Traditional numerical methods exhibit several inherent limitations, prompting Deep Learning (DL) as a promising alternative.
The effectiveness of current DL techniques typically depends on the availability of high-quantity and high-accuracy datasets, which are yet difficult to obtain in large deformation problems.
During the dataset construction process, a dilemma stands between data quantity and data accuracy, leading to suboptimal performance in the DL models.
To address this challenge, we focus on a representative application of large deformations, the stretch bending problem, and propose \textbf{FilDeep}, a \underline{\textbf{Fi}}delity-based Deep Learning framework for \underline{\textbf{l}}arge \underline{\textbf{De}}formation of \underline{\textbf{e}}lastic-\underline{\textbf{p}}lastic solids.
Our FilDeep aims to resolve the quantity-accuracy dilemma by simultaneously training with both low-fidelity and high-fidelity data, where the former provides greater quantity but lower accuracy, while the latter offers higher accuracy but in less quantity.
In FilDeep, we provide meticulous designs for the practical large deformation problem.
Particularly, we propose attention-enabled cross-fidelity modules to effectively capture long-range physical interactions across MF data.
To the best of our knowledge, our FilDeep presents the first DL framework for large deformation problems using MF data.
Extensive experiments demonstrate that our FilDeep consistently achieves state-of-the-art performance and can be efficiently deployed in manufacturing.
\end{abstract}




\maketitle

\section{Introduction}
In continuum mechanics, the scientific computation of large deformations in elastic-plastic solids is crucial for various manufacturing applications~\cite{deformation,application}.
The deformations of solids are typically induced by external loads and constraints~\cite{bending}.
Large deformations occur when the magnitude of deformations is substantial enough to invalidate the assumptions of infinitesimal strain theory~\cite{theory3}.
In such cases, the material undergoes a complex elastic-plastic process, where the elastic component of the deformation recovers upon unloading, while the plastic deformation remains permanent~\cite{bend2}.

\begin{figure}[t]
\setlength{\abovecaptionskip}{0.05cm}   %
    \centering
    \includegraphics[width=1\linewidth]{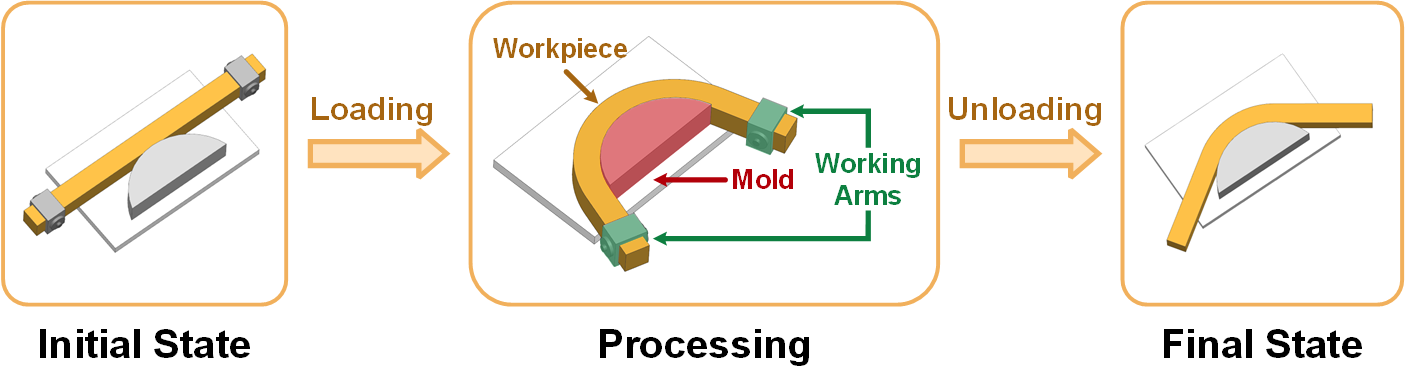}
    \caption{Stretch bending.}
    \label{fig:fig1}
    \vspace{-0.55cm} 
\end{figure}

Stretch bending is one of the most popular metal fabrication techniques involving large deformations~\cite{bend2}, as shown in Figure \ref{fig:fig1}.
In this process, the metal workpiece is positioned on the machine and securely clamped by two working arms.
These arms move and rotate (i.e., loading), exerting force on the workpiece, causing it to stretch and bend around a constrained mold.
Finally, the applied loads are released (i.e., unloading), allowing the workpiece to achieve its final shape with springback.
Given certain external conditions such as applied loads and motion boundaries, our primary interest is to predict the final shape of the workpiece~\cite{bending}.

Typically, the large deformation processes are considered to be described by Partial Differential Equations (PDEs)~\cite{theory3}.
Due to the intricate physical processes with unknown constitutive equations, their global PDEs are universally acknowledged as hard to formulate explicitly~\cite{theory2}.
Thus, traditional methods, e.g., the Finite Element Method (FEM)~\cite{FEM}, rely on local discrete linear approximations to obtain numerical solutions.
Yet in applications, these traditional methods often suffer from several inherent limitations, such as their poor trade-off between accuracy and efficiency, and their costly recalculation from scratch for every new instance~\cite{Transolver}.

Recently, Deep Learning (DL) has emerged as an effective tool for scientific and engineering problems, offering the potential to overcome the limitations of traditional numerical methods~\cite{ladeep, AI4S2}.
In practice, however, its effectiveness heavily depends on the availability of high-quantity and high-accuracy datasets~\cite{DL-data-hungry,MF-PINN}.
In many practical scenarios of science and engineering (e.g., the aforementioned stretch bending), such ideal datasets are often unattainable due to the high cost for data collection~\cite{MF-PINN,JSAC,TSC,TMC,kejia1,kejia2}.
As a result, the demand for data represents the most significant bottleneck and poses a fundamental challenge to leveraging DL techniques for modeling the large deformations of elastic-plastic solids.

Specifically, the common practice is to generate datasets through the aforementioned traditional numerical simulations~\cite{MF-FNO, MF-DeepONet1}. 
On the one hand, due to the extremely low efficiency of High-Fidelity (HF) simulations, we are often limited to obtaining few accurate data points, referred to as HF data~\cite{MF-review}.
Training DL models with few HF data may tend to risk overfitting with high variance~\cite{bias–variance}.
On the other hand, while we can actively accelerate the simulation process by introducing Low-Fidelity (LF) simplifications (e.g., coarser meshes), this comes at the expense of accuracy~\cite{MF-review}.
Although abundant, such LF data injects significant bias into the DL model, ultimately leading to degraded performance as well~\cite{bias–variance}.

Based on the above analyses, an inevitable dilemma appears to stand between data quantity and accuracy.
HF data can offer high accuracy but lack quantity, while LF data exhibit the opposite characteristic.
In this context, training with single-fidelity data will lead to suboptimal performance of the DL models, with either high variance or high bias~\cite{bias–variance,MF-PINN,MF-DeepONet0}.
Then a natural and interesting question arises: can we take the opportunity to generate Multi-Fidelity (MF) datasets instead of relying on a single fidelity level, thereby achieving both quantity and accuracy to overcome this dilemma?
In this way, we hope to simultaneously harness the strengths of both fidelity levels, i.e., the accuracy of HF data and the quantity of LF data, to construct an accurate and robust surrogate model.

To achieve this goal, in this paper, we propose a \underline{\textbf{Fi}}delity-based DL framework for \underline{\textbf{l}}arge \underline{\textbf{De}}formations of \underline{\textbf{e}}lastic-\underline{\textbf{p}}lastic solids, named \textbf{FilDeep}, as illustrated in Figure~\ref{fig:fig2}.
In our FilDeep, we first devise representations for the large deformation problem that not only adequately express the practical problem but are also suitable for DL.
Next, inspired by the MF simulation processes, we established an ``encoder-processor-decoder” structure for FilDeep, where the shared modules, input augmentation, and residual connection are employed to alleviate the challenge of HF data sparsity.
Considering that the deformations and contacts of metals are neither predetermined nor consistent across different fidelity levels, capturing the complex MF relationship is highly challenging.
To address this, we further propose Attention-enabled Cross-Fidelity (ACF) modules within the HF processor to account for the long-range nonlinear dependencies between the workpiece and external conditions across MF data.
Our main contributions are summarized as follows:
\begin{itemize}
    \item [(1)] Conceptually, we revealed the general quantity-accuracy dilemma for industrial dataset generation. To resolve it, our FilDeep is the first DL framework to leverage MF data for large deformation problems of elastic-plastic solids.
    \item [(2)] Technically, we developed a novel ACF to capture the long-range relationships across MF data for external-condition-aware modeling. To our knowledge, our FilDeep is the first work to leverage attention mechanisms in MF learning.
    \item [(3)] We contribute the first-ever MF dataset for large deformation problems to the community. This dataset was constructed using over 25,000 CPU hours, with all parameters derived from real-world applications.
    \item [(4)] Through extensive experiments, our FilDeep demonstrates superior performance compared to baseline methods. 
    Moreover, FilDeep has been successfully deployed in a real-world manufacturing factory, where its efficient inference and robust accuracy make it highly competitive.
\end{itemize}

\section{Background}
\label{motivation}
\subsection{Large Deformations of Elastic-Plastic Solids}
\label{no-PDE}
In continuum mechanics, deformation refers to the alteration in the shape or size of a solid object, which is typically induced by loading forces and motion boundaries~\cite{deformation}.
When deformations are very small and satisfy the infinitesimal strain theory, they can be described by simple linear elastic models, e.g., Hooke's law~\cite{theory}.
Large deformations refer to scenarios where the strains become significant enough to invalidate the assumptions of infinitesimal strain theory~\cite{theory}.
In this paper, we mainly focus on one of the most prevalent techniques involving large deformations: the stretch bending of metal~\cite{bend3}, as shown in Figure~\ref{fig:fig1}.
In this technique, metals are stretched and bent to achieve the desired shapes~\cite{bending}.

Typically, the whole process of a large deformation is considered to be governed by a set of nonlinear PDEs~\cite{theory3}.
This implies that the final steady state can be predetermined as long as the material properties, boundary conditions, and loads are fully provided.
Unfortunately, due to the unknown material constitutive equations, irregular solution domains, and unpredictable multi-object contacts, the explicit forms of these PDEs are typically impossible to write~\cite{theory2}.
To date, a general solution to such problems has remained elusive for more than half a century~\cite{theory2}.
See Appendix~\ref{appendix-deformation} for more details.

\begin{figure}[t]
\setlength{\abovecaptionskip}{0.2cm}   %
    \centering
    \includegraphics[width=0.90\linewidth]{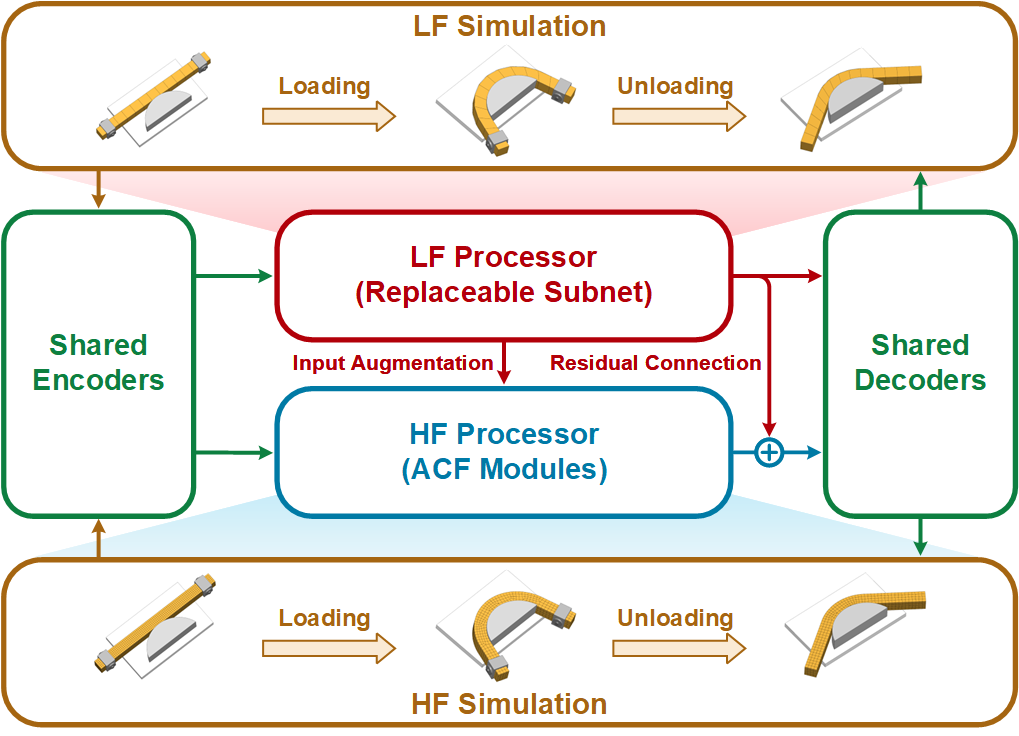}
    \caption{An overview of FilDeep.}
    \label{fig:fig2}
    \vspace{-0.5cm} 
\end{figure}

\subsection{Traditional Numerical Methods}
\label{tradition}
Due to the extreme cost of field experiments, traditional methods use numerical computations to approximately simulate the physical process, represented by the FEM~\cite{FEM}.
In FEM, the original materials are spatially partitioned into many small elements using meshing techniques. 
As for the time domain, the process of the large deformation is approached in a step-by-step manner, dividing the overall deformation into a series of small and manageable increments.
In each time step, the changes in deformation are assumed to be small enough to satisfy the infinitesimal strain theory~\cite{theory}.

In this way, the effectiveness and efficiency of FEM are highly dependent on the discretization of space and time.
For any computational instance, there exists a fundamental trade-off between computational accuracy and efficiency~\cite{Transolver}. Achieving higher accuracy necessitates a substantial increase in computational cost.
The computation has to be executed sequentially, 
and its significant time overhead can't be mitigated by parallel computing.
Additionally, the numerical computations must be restarted from scratch for every new instance, further exacerbating efficiency concerns in practical applications~\cite{Transolver}.
See Appendix~\ref{appendix-FEM0} for more details.

\subsection{Deep Learning Methods}
\label{background:DL}
Recently, DL has demonstrated significant potential in solving complex physical problems due to its capabilities for nonlinear representation~\cite{AI4S3, AI4S2, unisoma}. 
Current DL methods can be broadly categorized into two paradigms: physics-driven and data-driven approaches.
For the physics-driven paradigm, Physics-Informed Neural Networks (PINNs) stand out as representatives, which embed PDEs into the loss functions to approximate their solutions~\cite{PINNs, MF-PINN, physicsAI}. 
Unfortunately, such physics-driven methods require explicit PDE formulations and are completely inapplicable for complex problems without known PDEs, e.g., our large deformation problem~\cite{Transolver}.

As a result, we have to approximate the input-output mappings for our problem in a data-driven manner~\cite{FNO,DeepONet,SFNO}. 
However, its effectiveness typically depends on the availability of high-quantity and high-accuracy datasets~\cite{DL-data-hungry}, which are usually unattainable in practical industrial manufacturing scenarios.
Consequently, data availability emerges as the primary bottleneck for applying current DL techniques to solve large deformation problems of elastic-plastic solids.
In the following subsection, we will further elaborate on the specific quantity-accuracy dilemma within the data bottleneck.

\subsection{Quantity-Accuracy Dilemma}
\label{principle-dilemma}
To train DL models for physical problems, the common practice involves generating datasets through traditional simulations~\cite{MF-FNO, MF-DeepONet1}.
During this process, we identify an interesting and general Efficiency-Quantity-Accuracy (EQA) principle, revealing the triangular trade-off relationship in data construction.
It states that \textbf{high efficiency, large quantity, and precise accuracy, as three desirable characteristics for industrial dataset generation, cannot be achieved simultaneously in any simulation system}.

In practical industrial scenarios, cost budgets are always limited to some extent, imposing stringent constraints on efficiency within the EQA principle.
Consequently, the remaining two aspects, i.e., data quantity and data accuracy, inevitably face a dilemma.
Specifically, achieving high data accuracy requires costly HF simulations, resulting in low data quantity.
Conversely, simulation acceleration by LF simulation simplifications can be used for high data quantity, but it comes at the expense of data accuracy.
As a result, the DL models typically become suboptimal, suffering from either high bias or high variance~\cite{bias–variance}.
See Appendix~\ref{appendix-EQA} for more details.

\section{Basic Ideas \& Technical Challenges}
\label{basic-idea}
Based on our previous analysis, relying solely on single-fidelity data tends to lead to suboptimal performance.
It raises an interesting basic idea: can we use MF datasets for joint learning instead of relying on a single fidelity level to overcome the quantity-accuracy dilemma?
In such MF datasets, the HF data holds the advantage of accuracy, and the LF data can offer the advantage of quantity~\cite{MF2}.

\textbf{\textit{Remark:}} Simply mixing LF and HF data for training DL models is not feasible.
This is because, for the same input, the LF and HF outputs differ, which tends to harm the stability and convergence of the DL model.
Thus, to truly achieve MF training, the latent relationships across MF data must be captured.

Fortunately, according to the \textit{Universal Approximation Theorem}, neural networks themselves offer an ideal solution~\cite{MLP}. 
This allows us to design an MF framework consisting of two main sub-nets: one leverages the quantity of LF data to provide robust representations, while the other utilizes the accuracy of HF data for calibration.

For a given input $x$, we denote the outputs from LF data and HF data as $\bm{y}_{\mathrm{L}}$ and $\bm{y}_{\mathrm{H}}$, respectively.
We first consider a function $\mathcal{F}_{L}$ mapping the input $\bm{x}$ to the LF output $\bm{y}_{\mathrm{L}}$, which can be approximated by a neural network $\mathcal{N}_{L}$ with parameters $\theta_{L}$:
\begin{equation}
     \bm{y}_{\mathrm{L}}=\mathcal{F}_{L}(\bm{x})=\mathcal{N}_{L}(\bm{x};\theta_{L}).
\end{equation}
Then, we consider another function $\mathcal{F}_{\mathrm{H}}$ to capture the correlation across different fidelity levels. 
Specifically, this function, in addition to $\bm{x}$, takes the LF result $\bm{y}_{\mathrm{L}}$ as input and maps it to the HF output $\bm{y}_{\mathrm{H}}$.
This HF function can also be approximated by another neural network $\mathcal{N}_{\mathrm{H}}$ with parameters $\theta_{\mathrm{H}}$:
\begin{equation}
    \begin{aligned}
        \bm{y}_{\mathrm{H}} 
        = \mathcal{F}_{\mathrm{H}}(\bm{x}, \bm{y}_{\mathrm{L}}) 
        = \mathcal{N}_{\mathrm{H}}\left(\bm{x}, 
        \mathcal{N}_{L}\left({\bm{x}};\theta_{L}\right);\theta_{\mathrm{H}}\right).
    \end{aligned}
\end{equation}

In practice, applying this basic idea to our large deformation problem still presents several technical challenges to be addressed:
\begin{enumerate}
    \item \textbf{The representation of the problem.}
    Different from traditional DL tasks (e.g., computer vision and natural language processing) with well-defined input and output formats, our unstructured physical problem demands an appropriate representation for DL.
    We need to define formatted input and output representations for our real-world industrial problem to make them applicable for DL training.
    \item \textbf{The design of the MF neural network.}
    Although the basic idea outlines the macroscopic structure of the networks to utilize MF data, the detailed model requires meticulous design to adapt to our practical problem.
    This is particularly challenging as the cross-fidelity relationship can be highly complex, while the HF data are very sparse.
    \item \textbf{The training of DL model with MF data.}
    To address practical large-deformation problems in the absence of open MF datasets or benchmarks, we need to tackle this challenge to enable model training.
    Moreover, the loss functions and optimization strategies must also be meticulously designed to accommodate the characteristics of MF data.
\end{enumerate}

\section{Our Proposed FilDeep Framework}
\label{method}

\subsection{Problem Representation and  Formulation}
\label{representation}
This paper primarily focuses on the stretch bending problem, one of the most widely used fabrication processes involving large deformations. Specifically, our interest lies in predicting the final shape of the metal workpiece under given external conditions.
We generally represent the entire input as $\bm{x} = (\bm{w}, \bm{e})$, where $\bm{w}$ denotes the workpiece representations and $\bm{e}$ represents the external conditions.
The expected output $\bm{y}$ is the workpiece's final shape after springback.
Here, we use $\bm{y}_{\mathrm{L}}$ and $\bm{y}_{\mathrm{H}}$ to represent the LF and HF outputs, respectively.
The FilDeep framework can intuitively be seen as a deep surrogate for the function $f: \bm{x} \rightarrow \bm{y}_{\mathrm{H}}$.

For the workpiece $\bm{w}$, its cross-sectional shape remains constant throughout the stretch bending process.
Thus, each 3D workpiece can be fully represented by $\bm{w}=(\bm{s},\bm{l})$ with a 2D cross-section and a characteristic line, where the latter refers to the curve along the length of the workpiece.
We denote the cross-section of the workpiece as an image format $\bm{s} \in \mathbb{R}^{1 \times H \times W}$, and the characteristic line as a point set $\bm{l} \in \mathbb{R}^{M \times 3}$ of $M$ points with 3 dimensions.

As for the external conditions $\bm{e}$, we consider two key external factors: the motion trajectory guided by the loading arm and the motion boundaries constrained by the mold.
The motion trajectory of the loading arm can be represented by parameters with 6 degrees of freedom as $\bm{p} \in \mathbb{R}^{1 \times 6}$, effectively characterizing the accumulated impact of the loading arm on the motion state of the workpiece.
The motion boundaries can be represented by the shape of the mold, which can be adequately featured by the mold's characteristic line $\bm{m} \in \mathbb{R}^{M \times 3}$ alone, as the mold's cross-section is designed to only support the workpiece with no additional information.

In summary, the initial workpiece is represented by $\bm{w} = (\bm{s}, \bm{l})$, with its cross-section $\bm{s}$ and initial-state characteristic line $\bm{l}$.
The external conditions are represented by $\bm{e} = (\bm{p}, \bm{m})$, consisting of the motion parameters $\bm{p}$ and the mold's characteristic line $\bm{m}$.
Thus, the input $\bm{x} = (\bm{w}, \bm{e})$ can be expanded as $\bm{x} = (\bm{s}, \bm{l}, \bm{p}, \bm{m})$.
The output $\bm{y}\in \mathbb{R}^{M \times 3}$ is the workpiece's final-state characteristic line.

\subsection{Framework Overview}
\label{overview}
The overview of FilDeep is shown in Figure~\ref{fig:fig2}.
Inspired by the simulation processes of stretch bending, our FilDeep adopts a three-phase architecture with encoders, processors, and decoders:
\begin{itemize}
    \item Each simulation process begins with a start-up phase that involves creating components, assigning attributes, assembling parts, and defining boundary conditions and loads. This phase is represented by the encoders of FilDeep.
    \item Next, during the execution of simulations, variations in simulation settings (such as mesh division) lead to different stress distributions in the workpiece at different fidelity levels.
    Accounting for this, we introduce two different processors corresponding to the MF simulations.
    \item Finally, the decoders map the high-dimensional features from the processors back to the original space to predict the final shape of the workpiece after springback.
\end{itemize}

It is worth noting that, for the same problem instance, the start-up phase remains highly consistent across both LF and HF simulations, such as identical workpiece attributes, boundary conditions, and loads.
Motivated by this consistency, we designed shared encoders and decoders for both the LF and HF processors.
This shared design not only aligns with the physical consistency of the simulation but also significantly enhances model training. 
Specifically, our shared design can leverage the rich LF data to train robust encoders and decoders for the HF processor.
Without this shared design, training solely on the limited HF data may be very hard.

As for the processors, a key distinction between the HF processor and the LF processor lies in the HF processor’s ability to additionally read the output result from the LF processor to extract LF information.
This serves as an input enhancement for HF training, leveraging the feature enriched with LF information to facilitate easier training for the HF processor.
The rationale is intuitive: the HF processor can focus on learning HF outputs on the basis of LF output, rather than fitting the HF output from scratch.
Moreover, we also introduce a residual connection from the LF processor’s output to the HF processor’s output.
Thus, the HF processor only needs to focus on the gap between LF and HF, significantly reducing the burden compared to learning the HF output from scratch.

\textit{\textbf{Remark:}} 
{The scarcity of HF data poses the greatest bottleneck. 
Considering this, our FilDeep framework is meticulously designed with the shared modules, input augmentation, and residual connection.
These designs are primarily aimed at alleviating the learning burden on the HF processor and facilitating easier training for it.}

\begin{figure*}
    \centering
    \setlength{\abovecaptionskip}{0.05cm}   %
    \includegraphics[width=1\linewidth]{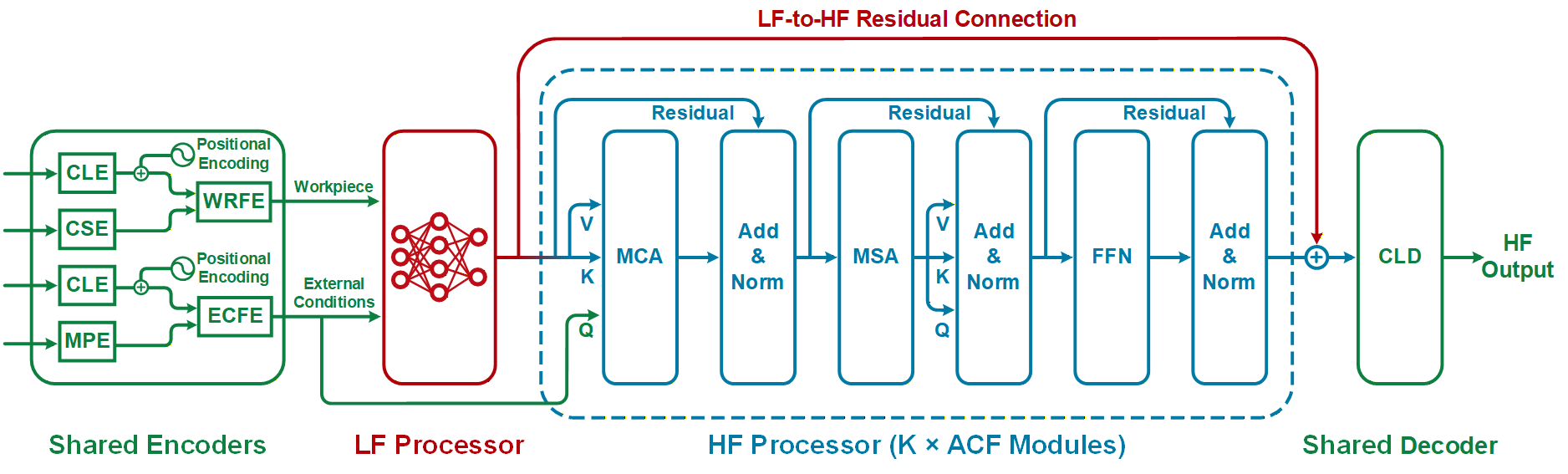}
    \caption{The forward process for HF training: First, the inputs are processed through the shared encoders to obtain a fused representation of the workpiece and external conditions, which are passed through the LF processor to obtain the LF workpiece representation $\tilde{\bm{w}}_{\mathrm{L}}$.
    The LF workpiece representation is then fed into the HF processor, which consists of $K$ stacked ACF modules.
    Finally, an additional LF-to-HF residual connection is incorporated into the HF processor’s output, and the high-dimensional feature is mapped back to the original space via CLD for the final output (i.e., the characteristic line of the final-state workpiece).
    }
    \label{fig:fig3}
    \vspace{-0.3cm} 
\end{figure*}

\subsection{Shared Encoders}
\label{encoders}

Based on our problem representation in Section \ref{representation}, we designed multiple shared encoders as follows.
After processing through all encoders, the representations are fused into two main features for the workpiece and external conditions, as shown in Figure~\ref{fig:fig3}.

\textbf{Characteristic Line Encoder (CLE).}
We first embed each point with an embedding size of $C$ using a linear layer.
To capture local curvature features, the points are partitioned into $N$ consecutive local regions and subsequently projected into tokens of $\mathbb{R}^{N \times C}$ by $N$ distinct linear layers.
Moreover, to incorporate global information, we use two linear layers to embed and encode the original points into a global feature of $\mathbb{R}^{1 \times C}$, which is repeated $N$ times and added to the local features.
To enhance the spatial position information within each characteristic line, positional encoding is incorporated into the final tokens.
Two distinct CLEs are used for the workpiece and the mold, producing outputs $\tilde{\bm{l}}, \tilde{\bm{m}} \in \mathbb{R}^{N \times C}$.

\textbf{Cross-Section Encoder (CSE).}
To provide a low-redundancy representation for cross-section, we adopt the Signed Distance Function (SDF) to omit unnecessary brightness, spectrum, and semantic information.
After getting the SDF representation,
we employ a pre-trained ResNet~\cite{restnet} as a frozen backbone with several additional trainable convolutional layers to extract features. 
Finally, the CSE results in a flattened output feature $\tilde{\bm{s}} \in \mathbb{R}^{1\times C}$.

\textbf{Motion Parameter Encoder (MPE).}
To comprehensively capture the influence of the movement effect along each degree of freedom, we first embed each parameter with a size of $C$.
Subsequently, these parameters are individually projected into $F$ tokens using 6 distinct linear layers, resulting in a total of $N = F \times 6$ tokens.
Thus, the final output of MPE is tokens $\tilde{\bm{p}} \in \mathbb{R}^{N \times C}$.

\textbf{Workpiece Representation Fusion Encoder (WRFE).}
As described above, the workpiece is represented by two separated encoded features, the cross-section $\tilde{\bm{s}} \in \mathbb{R}^{1\times C}$ and the characteristic line $\tilde{\bm{l}} \in \mathbb{R}^{N \times C}$.
Here, our WRFE is designed to combine these two features for a comprehensive workpiece representation.
Specifically, we repeat the cross-section $\tilde{\bm{s}}$ by $N$ times and add it into $\tilde{\bm{l}}$ to get the fused workpiece feature $\tilde{\bm{w}} \in \mathbb{R}^{N \times C}$.

\textbf{External Condition Fusion Encoder (ECFE).}
Let's recall that the mold motion parameters $\tilde{\bm{p}} \in \mathbb{R}^{N \times C}$ and the mold’s characteristic line $\tilde{\bm{m}} \in \mathbb{R}^{N \times C}$ together constitute the external conditions for the stretch bending process.
To obtain a unified representation, we concatenate these two features and map them to the fused external condition feature $\tilde{\bm{e}} \in \mathbb{R}^{N \times C}$ using a linear layer.

\subsection{Multi-Fidelity Processors}
As shown in Figure~\ref{fig:fig2}, the MF processors of FilDeep consist of two separate subnets for both LF and HF.
Recall that the LF data are abundant and don't pose a major bottleneck.
Thus, we focus on the HF processor and design Attention-enabled Cross-Fidelity (ACF) modules for it, extracting the long-range cross-fidelity relationships.
To the best of our knowledge, our FilDeep represents the first work to leverage attention mechanisms in MF learning.

\textbf{Why Attention?}
As outlined in Section \ref{overview}, input augmentation and residual connection are utilized for the HF processor.
\textbf{This requires the HF processor to capture the complex relationships across different fidelity levels, involving long-range nonlinear dependencies between the workpiece and the external conditions.}
For example, the final shape of the workpiece is significantly influenced by the clamp's motion parameters.
These parameters have a global effect on the workpiece, as the motion at the clamped end induces internal stresses that propagate through the workpiece material.
These stresses will lead to deformations in regions not directly in contact with the clamps.
Similarly, the mold as the boundary also exerts global influences on the workpiece, rather than being limited to the local one at their contact points.

The specific forward process for HF training of our FilDeep is shown in Figure~\ref{fig:fig3}.
Our HF processor is designed as a composition of $K$ stacked ACF modules, which are sequentially constructed using a Multi-head Cross-Attention (MCA), a Multi-head Self-Attention (MSA), and a Feed-Forward Network (FFN).
In the MCA, we use the fused external conditions as the query and the LF-processed workpiece as the key and value, aiming to capture the global effects of external conditions on the workpiece across different fidelity levels.
Subsequently, the MSA takes the processed workpiece as the query, key, and value to capture the intricate relationships within the LF workpiece.
Finally, the FFN is incorporated to enhance nonlinear modeling capabilities.
We employ a residual connection~\cite{restnet} around each of the two sub-layers, followed by layer normalization~\cite{layer-norm}.
Furthermore, an LF-to-HF residual connection is introduced from the LF processor's output to the HF processor, encouraging the ACF modules to learn the gap between fidelities.

We use the feature $\tilde{\bm{e}} \in \mathbb{R}^{N \times C}$ to represent the fused external conditions, and denote the LF workpiece representation (i.e., the result of the LF processor) as $\tilde{\bm{w}}_{\mathrm{L}} \in \mathbb{R}^{N \times C}$. 
We set the initial input of the HF processor, $\hat{\bm{w}}_{\mathrm{H}}^{(0)}$, to be the LF workpiece representation $\tilde{\bm{w}}_{\mathrm{L}}$.
The general workflow of our HF processor consists of a total of $K$ stacked ACF modules, as represented below.
\begin{align*}
\label{ACF}
{\scriptstyle
k\in[1, \, K]}
\left\{
\begin{aligned}
    &\hat{\bm{w}}_\mathrm{cross}^{k}=
    \mathrm{LN}\left(
        \mathrm{MCA}\left(
    \hat{\bm{e}}_{\mathrm{H}},\hat{\bm{w}}_{\mathrm{H}}^{k-1},\hat{\bm{w}}_{\mathrm{H}}^{k-1}
\right)
    + \hat{\bm{w}}_{\mathrm{H}}^{k-1}
    \right), \\
    &\hat{\bm{w}}_{\mathrm{self}}^{k}=
    \mathrm{LN}\left(
    \mathrm{MSA}\left( \hat{\bm{w}}_{\mathrm{cross}}^{k},\hat{\bm{w}}_{\mathrm{cross}}^{k},\hat{\bm{w}}_{\mathrm{cross}}^{k}
    \right) 
    + \hat{\bm{w}}_{\mathrm{cross}}^{k}
    \right), \\
    &\hat{\bm{w}}_{\mathrm{H}}^{k}=
\mathrm{LN}\left(
    \mathrm{FFN}\left(\hat{\bm{w}}_\mathrm{self}^{k}\right) + \hat{\bm{w}}_\mathrm{self}^{k}
\right).
\end{aligned}
\right.
\end{align*}

Here, $\mathrm{LN}(\cdot)$ denotes layer normalization. 
The $\mathrm{MCA}(\mathcal{Q,K,V})$ and the $\mathrm{MSA}(\mathcal{Q,K,V})$ can utilize multi-head structures for attending to information from diverse representation subspaces.

\textit{\textbf{Remark:}} 
One key difference between the HF processor and the LF one is that, the former does not utilize the raw workpiece feature $\tilde{\bm{w}}$. 
Instead, it leverages the LF-processed feature 
$\hat{\bm{w}}_{\mathrm{L}}$ as an enhanced substitute, which further integrates additional LF information.

\subsection{Shared Decoders and Loss Functions}
\label{decoder}
After obtaining the high-dimensional representations from the MF processors, we design two shared decoders to map these representations back to the original space for the final output.

\textbf{Cross-Section Decoder (CSD).}
To reconstruct the cross-section, we adopt the structure of a Variational Auto Encoder (VAE) to devise our CSD, using deconvolution layers with interpolation operations.

\textbf{Characteristic Line Decoder (CLD).}
To decode the representation of the characteristic line back to the original space, we construct the CLD by approximately reversing the structure of the CLE.

Based on the results obtained from the above decoders, we further introduce two distinct types of loss functions as follows.

\textbf{Cross-Section Loss (CSL).}
The CSL is used to extract informative features for the cross-section by reconstruction.
Specifically, the Mean Squared Error (MSE) is computed between the output of the CSD and the ground truth SDF of the workpiece's cross-section. 

\textbf{Characteristic Line Loss (CLL).}
The CLL is used to guide the prediction of the workpieces' characteristic lines.
Considering the imbalanced value distributions across different coordinate axes of the characteristic line with 3 dimensions, normalizing all axes equally could lead to inaccurate shifts.
Thus, we employ a coordinated $L_{2}$ loss to emphasize axes with larger orders of magnitude.

\begin{table*}[!ht]
\centering
\caption{Performance comparison with baselines. Each baseline was trained separately using single LF data, single HF data, simply mixed MF data, and FilDeep-supported MF data.
Within FilDeep, each baseline functions as the LF processor.
For clarity, the best result is highlighted in bold and underlined, while the second-best result is marked with a wavy underline.
The Adv. refers to the relative reduction in MAD and TE, as well as the relative improvement in 3D IoU w.r.t. the second-best model.} 
\vspace{-3mm}
\resizebox{\textwidth}{!}{
\begin{NiceTabular}{@{}c|ccccc|ccccc|ccccc@{}}
\toprule
\multicolumn{1}{c|}{\multirow{2}{*}{{Baseline}}} &
  \multicolumn{5}{c|}{MAD (mm) {\bm{$\downarrow$}}} &
  \multicolumn{5}{c|}{3D IoU (\%) {\bm{$\uparrow$}}} &
  \multicolumn{5}{c}{TE (mm) {\bm{$\downarrow$}}} \\ \cmidrule(l){2-16} 
\multicolumn{1}{c|}{} &
  LF &
  HF &
  Mix &
  \multicolumn{1}{c|}{FilDeep} &
  {Adv.} &
  LF &
  HF &
  Mix &
  \multicolumn{1}{c|}{FilDeep} &
  {Adv.} &
  LF &
  HF &
  Mix &
  \multicolumn{1}{c|}{FilDeep} &
  {Adv.} \\ \midrule
Transformer~\cite{Attention}         & 11.13 & \uwave{1.61} & 11.66 & \multicolumn{1}{c|}{\underline{\textbf{0.50}}} & \textcolor{c2}{\textbf{69.01\%}} & 2.38 & \uwave{36.87} & 2.38 & \multicolumn{1}{c|}{\underline{\textbf{70.83}}}  & \textcolor{c2}{\textbf{92.10\%}} & 17.79 & \uwave{4.81} & 19.64 &\multicolumn{1}{c|}{\underline{\textbf{1.44}}} & \textcolor{c2}{\textbf{70.04\%}} \\
DeepONet~\cite{DeepONet}   & 10.61 & \uwave{3.97} & 11.67 & \multicolumn{1}{c|}{\underline{\textbf{0.56}}} & \textcolor{c2}{\textbf{86.90\%}} & 2.44 & \uwave{31.07} & 2.34 & \multicolumn{1}{c|}{\underline{\textbf{61.12}}}  & \textcolor{c2}{\textbf{96.72\%}} & 14.22 & \uwave{8.85} & 20.02 &\multicolumn{1}{c|}{\underline{\textbf{1.66}}} & \textcolor{c2}{\textbf{81.25\%}}  \\
IDeepONet~\cite{improved-deeponet}   & 10.33 & \uwave{3.60} & 11.23 & \multicolumn{1}{c|}{\underline{\textbf{0.69}}} & \textcolor{c2}{\textbf{80.77\%}} & 2.41 & \uwave{35.44} & 2.38 & \multicolumn{1}{c|}{\underline{\textbf{58.35}}}  & \textcolor{c2}{\textbf{64.65\%}} & 16.13 & \uwave{8.03} & 18.90 &\multicolumn{1}{c|}{\underline{\textbf{2.07}}} & \textcolor{c2}{\textbf{74.25\%}}  \\
FNO~\cite{FNO-2}        & 9.31 & \uwave{4.21} & 10.98 & \multicolumn{1}{c|}{\underline{\textbf{0.55}}} & \textcolor{c2}{\textbf{86.93\%}} & 2.75 & \uwave{33.98} & 2.62 & \multicolumn{1}{c|}{\underline{\textbf{66.43}}}  & \textcolor{c2}{\textbf{95.48\%}} & 13.93 & \uwave{9.55} & 18.87 &\multicolumn{1}{c|}{\underline{\textbf{1.63}}} & \textcolor{c2}{\textbf{82.89\%}}  \\
SFNO~\cite{SFNO}      & 10.48 & \uwave{2.56} & 11.23 & \multicolumn{1}{c|}{\underline{\textbf{0.92}}} & \textcolor{c2}{\textbf{64.21\%}} & 2.43 & \uwave{39.65} & 2.45 & \multicolumn{1}{c|}{\underline{\textbf{54.85}}}  & \textcolor{c2}{\textbf{38.31\%}} & 16.56 & \uwave{6.59} & 19.01 &\multicolumn{1}{c|}{\underline{\textbf{2.56}}} & \textcolor{c2}{\textbf{61.23\%}}  \\
TFNO~\cite{TFNO}        & 9.83 & \uwave{4.68} & 11.28 & \multicolumn{1}{c|}{\underline{\textbf{1.56}}} & \textcolor{c2}{\textbf{66.74\%}} & 2.50 & \uwave{31.84} & 2.39 & \multicolumn{1}{c|}{\underline{\textbf{41.72}}}  & \textcolor{c2}{\textbf{31.02\%}} & 14.81 & \uwave{9.58} & 19.38 &\multicolumn{1}{c|}{\underline{\textbf{3.06}}} & \textcolor{c2}{\textbf{68.03\%}}  \\
UNO~\cite{UNO}        & 10.28 & \uwave{2.05} & 11.34 & \multicolumn{1}{c|}{\underline{\textbf{0.60}}} & \textcolor{c2}{\textbf{70.56\%}} & 2.38 & \uwave{41.83} & 2.36 & \multicolumn{1}{c|}{\underline{\textbf{62.97}}}  & \textcolor{c2}{\textbf{50.55\%}} & 15.76 & \uwave{5.59} & 19.02 &\multicolumn{1}{c|}{\underline{\textbf{1.64}}} & \textcolor{c2}{\textbf{70.72\%}}  \\
\bottomrule
\end{NiceTabular}
}
\label{overall}
\end{table*}

\section{Experiments}
\label{experiments}
This section validates our FilDeep. 
We conducted extensive experiments to answer the following Research Questions (RQ):

\begin{itemize}
    \item \textbf{RQ1:} How does FilDeep outperform single-fidelity learning by leveraging MF data, and how does this capability vary when employing different LF processors?
    \item \textbf{RQ2:} What is the effect of the designed module and hyper-parameters on the performance of FilDeep?
    \item \textbf{RQ3:} How does FilDeep show its ability to capture physical features and interaction patterns in visualizations?
    \item \textbf{RQ4:} How does FilDeep outperform traditional numerical methods, and how can we deploy it in practical scenarios?
\end{itemize}

\subsection{Experiments Setup}
\subsubsection{Datasets}
Due to the absence of open MF datasets for our large deformation problems, we meticulously construct an MF dataset to support the training and evaluation of FilDeep.
In our dataset, all the input representations are derived from real-world applications, and the MF data are constructed separately based on MF simulations.
In this way, we spent a total computational cost exceeding 25,000 CPU hours to construct 3,000 LF instances and 300 HF instances.
This dataset will be open-sourced alongside the FilDeep code at  \url{https://github.com/tangent-heng/FilDeep}.

\subsubsection{Baselines}
\label{Baselines}
To evaluate our FilDeep's effectiveness in resolving the quantity-accuracy dilemma, we introduced a series of well-known DL baselines, including Transformer~\cite{Attention}, DeepONet~\cite{DeepONet}, IDeepONet~\cite{improved-deeponet}, FNO~\cite{FNO-2}, SFNO~\cite{SFNO}, TFNO~\cite{TFNO}, and UNO~\cite{UNO}.
These baselines not only serve as backbone networks for training single-fidelity and simple-mixed data, but also function as the LF processor within FilDeep.
Recall that the large deformations lack explicit PDE formulations, making PINN-based methods unsuitable as baselines.
Moreover, to show FilDeep's superiority in capturing MF relationships, we introduced several popular MF methods, including MFNN~\cite{MF-PINN}, MF-DeepONet~\cite{MF-DeepONet0}, and their residual variants~\cite{Extraction,MF-DeepONet0}.
Finally, considering the employment in the real-world manufacturing factories, we further compared FilDeep with the most widely used FEM~\cite{FEM} to highlight its practical efficiency.

\subsubsection{Metrics}
To evaluate inference accuracy (primarily for comparison with DL methods), we introduced three metrics to provide a comprehensive benchmark: Mean Absolute Distance (MAD), 3-Dimensional Intersection over Union (3D IoU), and Tail Error (TE).
The MAD measures the mean absolute error between the predicted workpiece’s characteristic line and the ground truth. 
The 3D IoU measures the overlap ratio between the reconstructed workpiece and the ground truth object in 3D.
The TE measures the mean absolute error between the predicted points on the tail surface and the ground truth, as errors tend to accumulate on the tail surface for large deformation predictions.
For inference efficiency, from the perspective of practical applications (primarily for comparison with FEM), we used inference time as the primary evaluation metric.

\subsection{Overall Evaluation (RQ1)}
\label{Overall Evaluation}

\begin{figure*}[h]
\setlength{\abovecaptionskip}{0.2cm}   %
	\begin{minipage}[t]{0.33\linewidth}
		\centering	\includegraphics[width=1.0\textwidth]{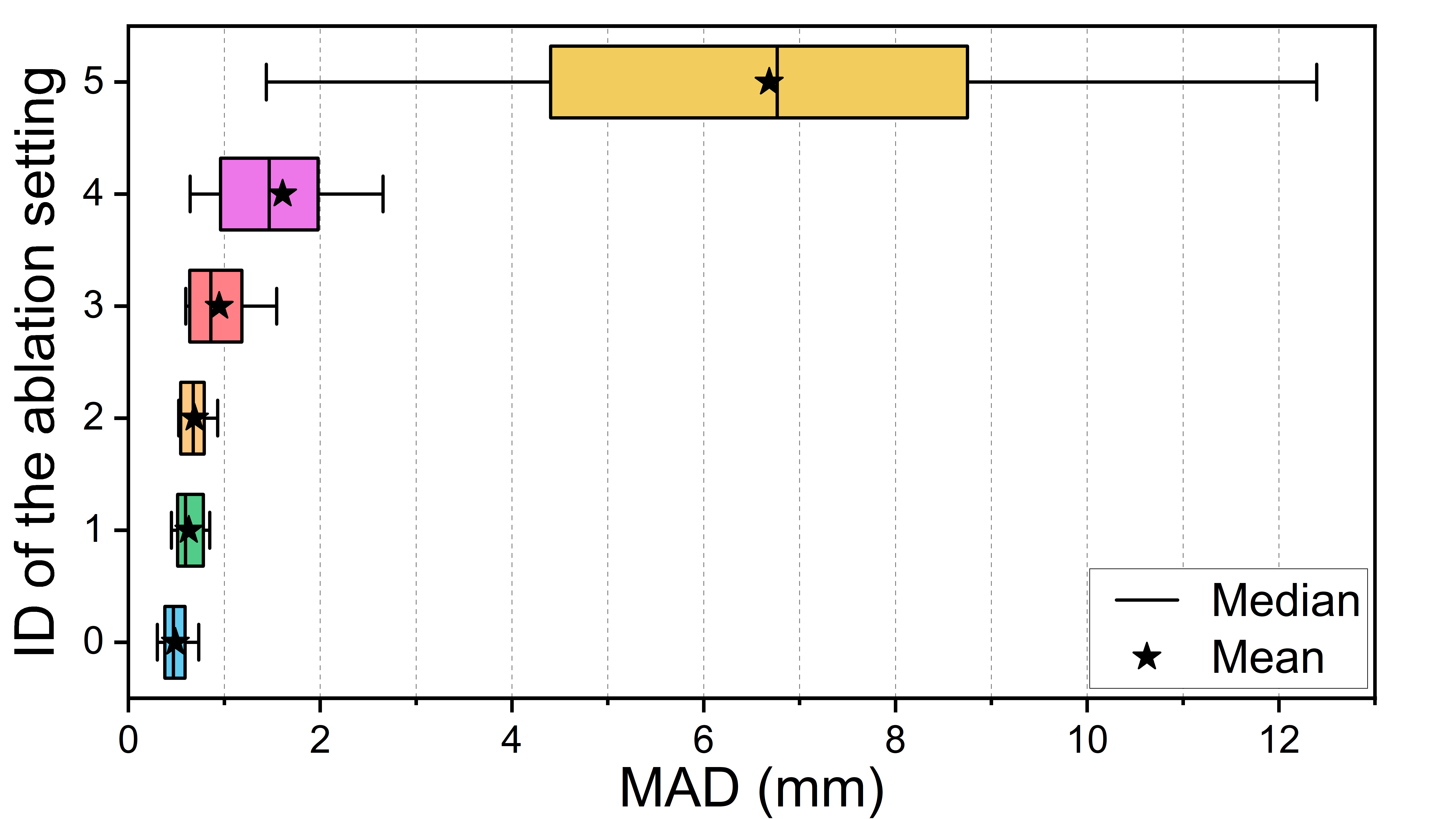}
        \caption{Structural ablation study.}
    \label{fig:ablation}
	\end{minipage}
	\hfill 
        \begin{minipage}[t]{0.33\linewidth}
		\centering
\includegraphics[width=1.0\textwidth]{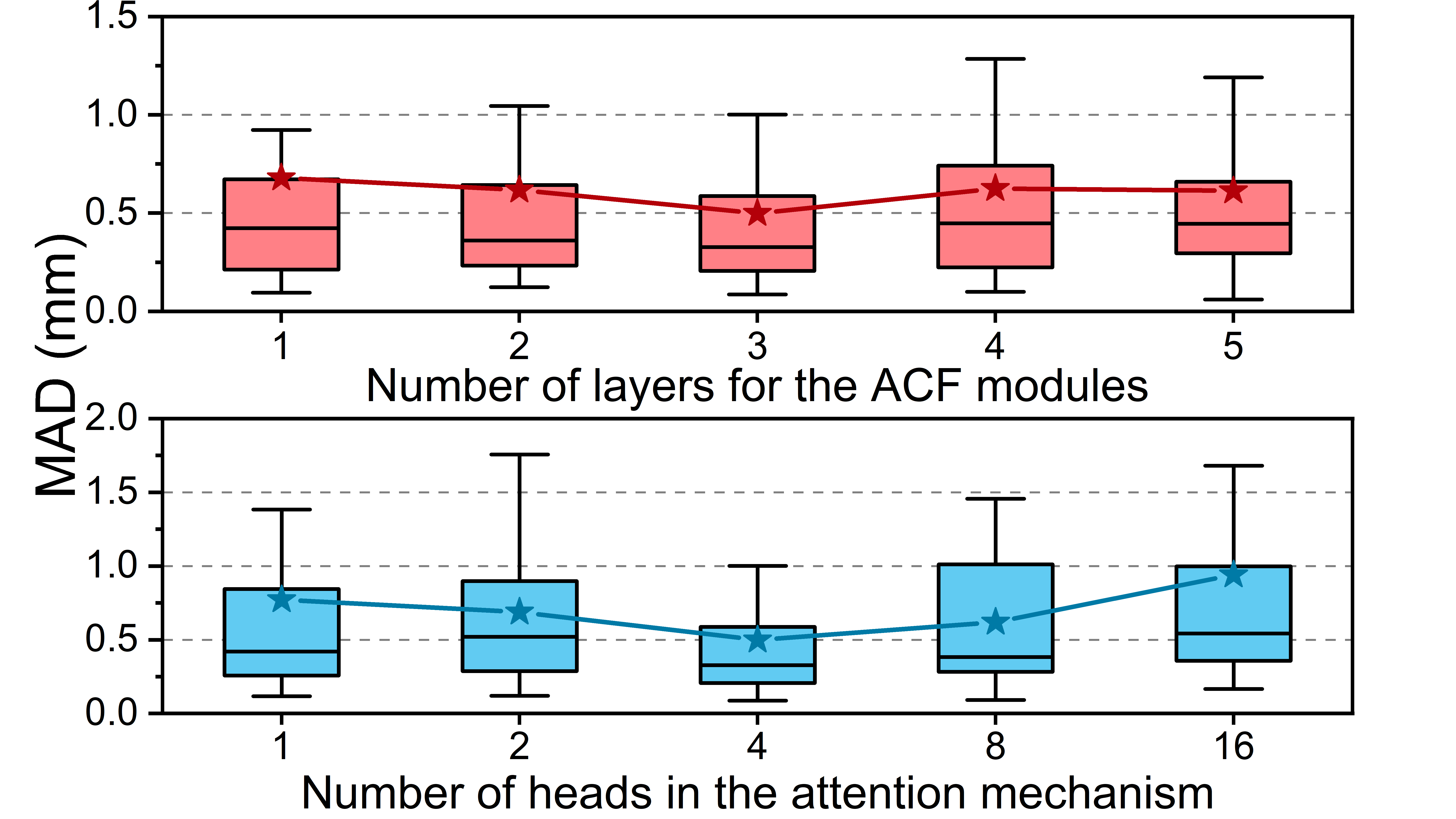}
        \caption{Hyper-parameters analysis.}
    \label{fig:params}
	\end{minipage}
	\hfill 
	\begin{minipage}[t]{0.33\linewidth}
		\centering	\includegraphics[width=1.0\textwidth]{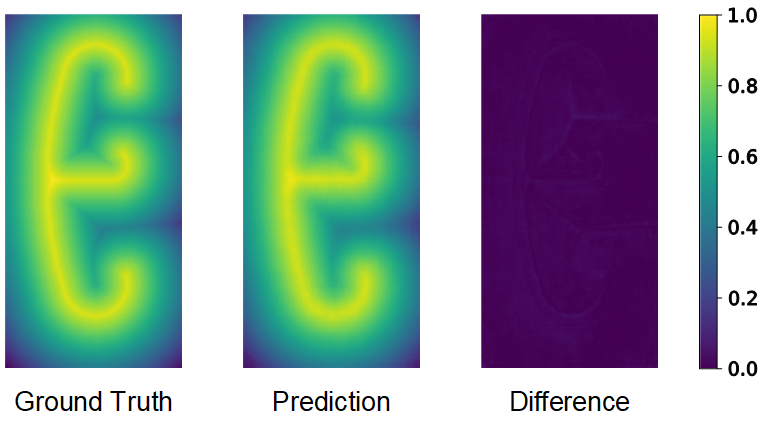}
        \caption{Cross-section reconstruction.}
    \label{fig:cross-section}
	\end{minipage}
    \vspace{-0.2cm} 
\end{figure*}

\begin{figure*}[h]
\setlength{\abovecaptionskip}{0.2cm}   %
	\begin{minipage}[t]{0.33\linewidth}
		\centering	\includegraphics[width=1.0\textwidth]{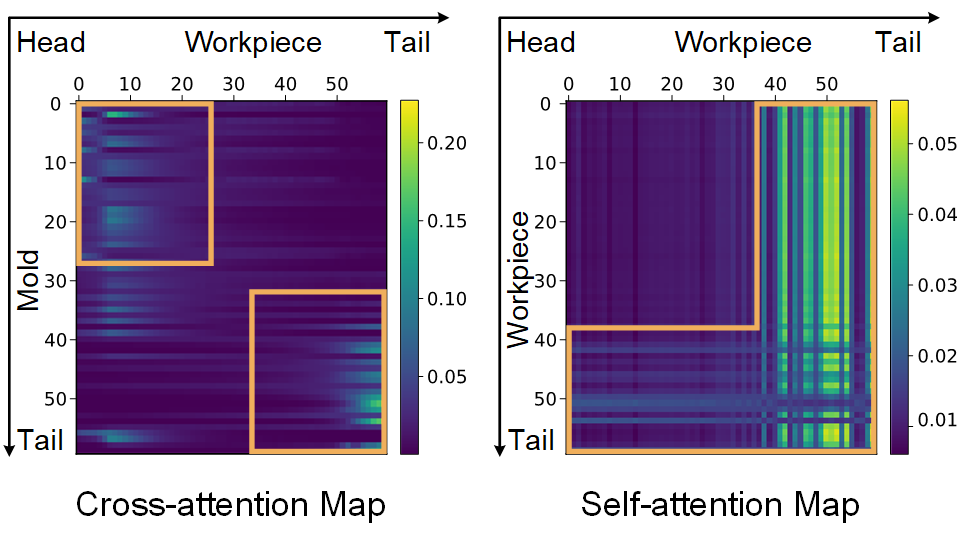}
        \caption{Attention map visualization.}
    \label{fig:attention}
	\end{minipage}
        \hfill
        \begin{minipage}[t]{0.33\linewidth}
		\centering
\includegraphics[width=1.0\textwidth]{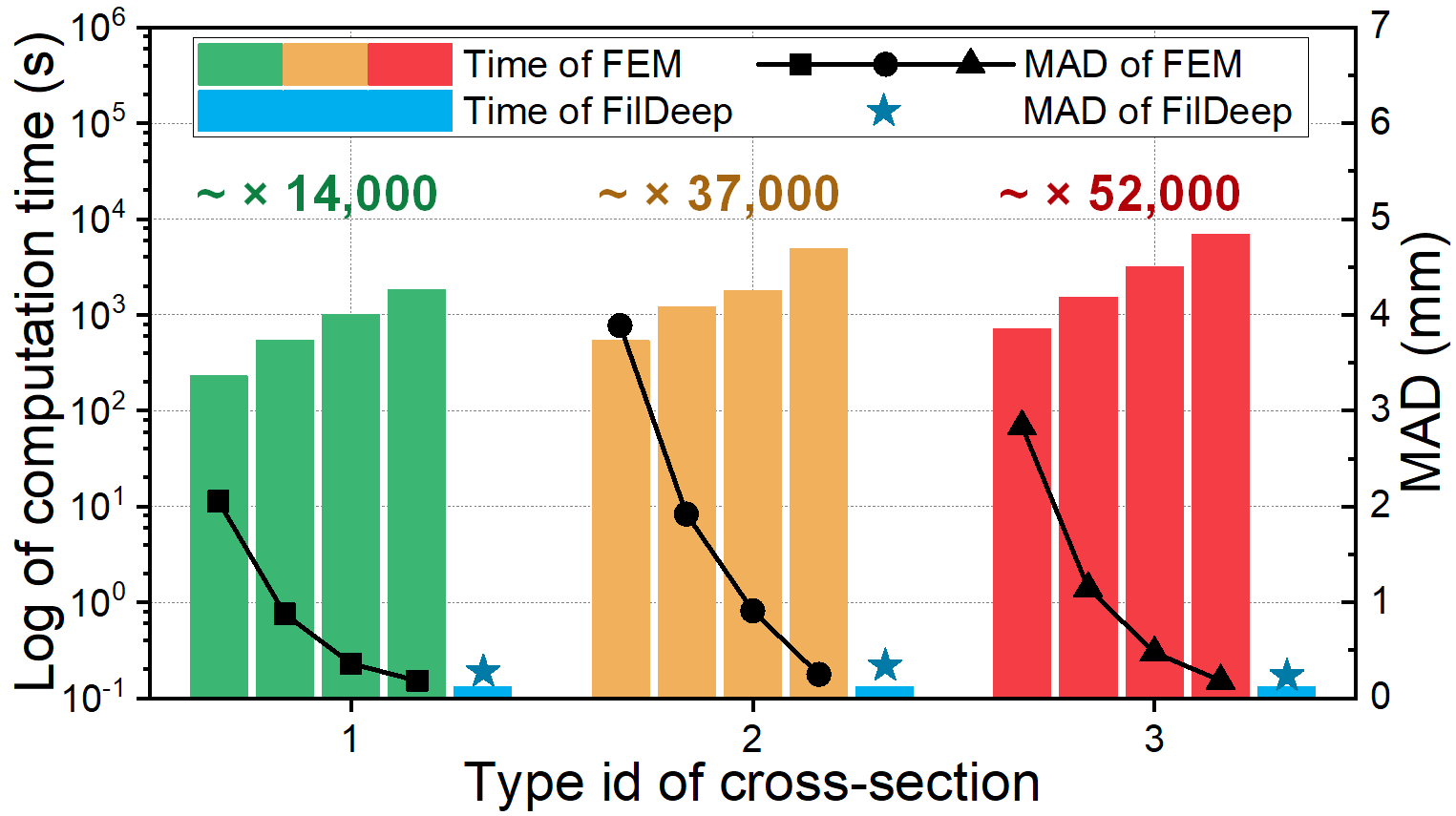}
		\caption{Comparison with FEM.}
    \label{fig:acc-eff-1}
	\end{minipage}
	\hfill 
	\begin{minipage}[t]{0.33\linewidth}
		\centering	\includegraphics[width=1.0\textwidth]{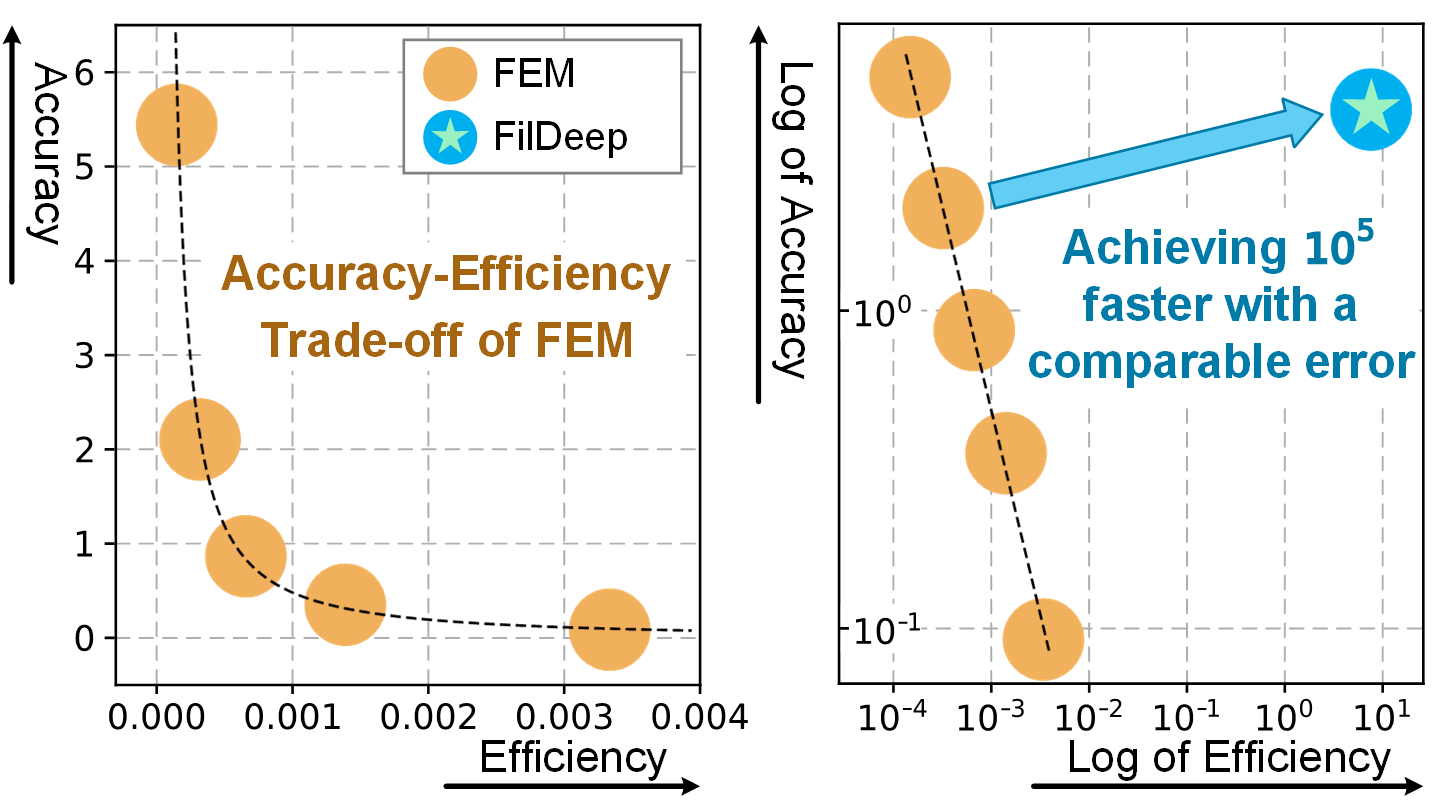}
		\caption{Overview of performance trends.}
    \label{fig:acc-eff-2}
	\end{minipage}
    \vspace{-0.2cm} 
\end{figure*}

To demonstrate the superiority of our FilDeep trained with MF data over training with only single-fidelity data, we conducted comprehensive overall evaluations with various DL baselines, which are reported in Table~\ref{overall}.
To ensure that all methods are well-suited to our problem and enable fair comparisons, we consistently used our designed encoders and decoders across all methods to extract effective features.
Meanwhile, to simulate scenarios where HF data are scarce, each model was trained using 60 HF samples, with another 30 HF samples allocated for validation and 30 HF samples allocated for testing.
As shown in the results, the models trained with single LF data exhibit the poorest performance across different baselines due to the inherent bias in the LF data.
An interesting finding is that FilDeep successfully extracts useful information from such noisy and biased LF data, achieving performance gains by capturing the LF-to-HF relationships.
As a result, FilDeep consistently outperforms other baselines and achieves state-of-the-art results.

\begin{table}[t]
\centering
\caption{Ablations of the ACF module with MF baselines}
\vspace{-3mm}
\label{tab:MF}
\resizebox{\columnwidth}{!}{%
\renewcommand{\arraystretch}{1.1}
\begin{NiceTabular}{llll}
\toprule

\textbf{Method for HF Processor}      & \textbf{MAD (mm)} {\bm{$\downarrow$}}    & \textbf{3D IoU} (\%) {\bm{$\uparrow$}}   & \textbf{TE (mm)} {\bm{$\downarrow$}}     \\ \hline
FilDeep    & 0.50 & 70.83 & 1.44  \\
FilDeep without Residual    & 0.65 (\textcolor{c1}{\textbf{\bm{$\uparrow$} 30.00\%}}) & 61.88 (\textcolor{c1}{\textbf{\bm{$\downarrow$} 12.64\%}}) & 1.84 (\textcolor{c1}{\textbf{\bm{$\uparrow$} 27.78\%}})  \\
Residual MFNN~\cite{Extraction}    & 0.75 (\textcolor{c1}{\textbf{\bm{$\uparrow$} 50.00\%}}) & 55.39 (\textcolor{c1}{\textbf{\bm{$\downarrow$} 21.80\%}}) & 2.17 (\textcolor{c1}{\textbf{\bm{$\uparrow$} 50.69\%}})  \\
Residual MF-DeepONet~\cite{MF-DeepONet0}    & 0.68 (\textcolor{c1}{\textbf{\bm{$\uparrow$} 36.00\%}}) & 59.78 (\textcolor{c1}{\textbf{\bm{$\downarrow$} 15.60\%}}) & 2.87 (\textcolor{c1}{\textbf{\bm{$\uparrow$} 99.31\%}})  \\ 
Vanilla MFNN~\cite{MF-PINN}    & 7.63 (\textcolor{c1}{\textbf{\bm{$\uparrow$} 1426\%}}) & 12.71 (\textcolor{c1}{\textbf{\bm{$\downarrow$} 82.06\%}}) & 19.56 (\textcolor{c1}{\textbf{\bm{$\uparrow$} 1258\%}})  \\
Vanilla MF-DeepONet~\cite{MF-DeepONet1}    & 7.60 (\textcolor{c1}{\textbf{\bm{$\uparrow$} 1420\%}}) & 13.29 (\textcolor{c1}{\textbf{\bm{$\downarrow$} 81.24\%}}) & 19.53 (\textcolor{c1}{\textbf{\bm{$\uparrow$} 1256\%}})  \\
\bottomrule
\end{NiceTabular}%
}
\vspace{-5mm}
\label{exp:abalation}
\end{table}

Notably, when using the Transformer as the backbone, it achieves the best performance within FilDeep, reaching 0.50 mm in MAD, 70.83\% in 3D IoU, and 1.44 mm in TE.
This evidence may suggest that the state-of-the-art performance of the Transformer within FilDeep is not solely attributed to its strong expressive capability but rather to the synergy between its attention-enabled design and the ACF module in our FilDeep.
In addition, it is important to highlight that simply mixing MF data provides little to no benefit and, in some cases, even leads to slight negative effects. 
This observation stems from the differences between LF and HF outputs for the same input.
Such differences can impede the stability and convergence of the DL model, consistent with our earlier analysis in Section~\ref{basic-idea}.

\newpage
\vspace{0.1cm}

\subsection{Ablation Study (RQ2)}
\label{ablation-study}
\vspace{0.1cm}

To verify the effectiveness of our designed module and the impact of hyper-parameters in FilDeep, we conducted extensive experiments for the ablation study.
Here, our ablation study consisted of three main parts: (1) analyzing the superiority of our ACF module in learning MF relationships, especially in comparison with other MF methods; (2) evaluating the effectiveness of each component design in FilDeep; and (3) performing hyper-parameter analyses.

To demonstrate the superiority of our FilDeep in learning MF relationships for our large deformation problem, we introduced several state-of-the-art MF baselines for comparison, including MFNN~\cite{MF-PINN}, MF-DeepONet~\cite{MF-DeepONet1}, and their residual variants~\cite{Extraction,MF-DeepONet0}.
Note that these baselines are not initially designed for large deformation problems and lack effective encoders and decoders.
Therefore, to ensure fair comparisons, we maintained the overall framework of FilDeep and replaced only the HF processor with the configurations of these MF baselines.
Additionally, we also performed an ablation study on the residual connection of our FilDeep to explore its robustness without the LF-to-HF connection.

As shown in Table~\ref{tab:MF}, the results demonstrate the superiority of our proposed ACF module, which is the first to apply the attention mechanism in MF learning.
This advantage arises because existing MF baselines were initially designed for simple and structured problems, relying solely on feature-level mappings.
What distinguishes our FilDeep is that it further accounts for the global influence of external conditions on the workpiece in complex large deformation scenarios.
It is emphasized that, in our large deformation problem, the baselines are highly sensitive to residuals and tend to fail or even collapse without residuals, indicating that they cannot effectively capture the MF relationship.
In contrast, our FilDeep shows much less performance degradation when the residual connection is removed, indicating that it is much less sensitive to residual connections.
Notably, even without the residual connection, our FilDeep still outperforms all other MF baselines, further validating the superiority of the attention mechanism in our ACF module for learning the MF relationship in our large deformation problem.

\begin{figure}[t]
    \centering
    \includegraphics[width=0.45\textwidth]{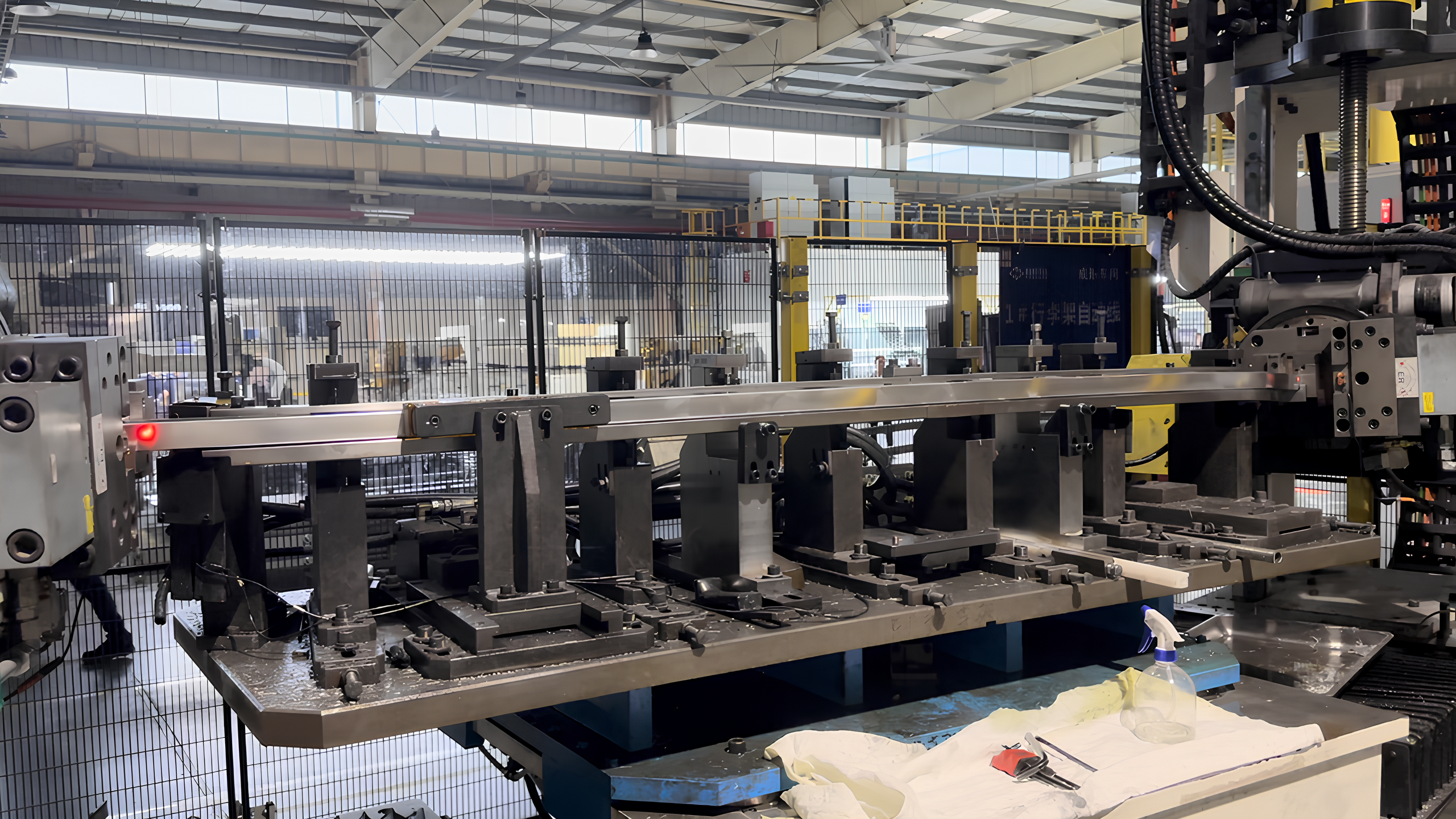}
		\caption{A real mold for on-site production.}
    \label{fig:appendix-mold-new}
    \vspace{-5mm}
\end{figure}

Next, to conduct comprehensive structural ablation experiments, we built several distinct variants of FilDeep.
The variants for structural ablation included (with index 0 as the non-ablated FilDeep):
(1) w/o LF-to-HF residual connection,
(2) w/o shared parameters in only the decoders,
(3) w/o shared parameters in both the encoders and decoders,
(4) replacing CLL with MSE, which treats all axes equally, and (5) replacing encoders and decoders with MLPs.
The structural ablation results are shown in Figure~\ref{fig:ablation}, showing the necessity of these designs in FilDeep.
Among them, the encoders and decoders have the greatest impact on the performance of FilDeep, highlighting the effectiveness of our tailored designs.

Finally, we conducted ablation studies on hyper-parameters, as shown in Figure~\ref{fig:params}.
Specifically, we focused on exploring the impact of the number of layers in the ACF module and the number of heads in its attention mechanism on the performance of FilDeep.
We can observe that as the number of ACF module layers increases, the MAD generally decreases first and then increases, and the optimal number of ACF modules is 3.
This is very intuitive, as too many ACF layers introduce a large number of model parameters, which is detrimental to the sparse HF data.
In addition, as the number of attention heads increases, the MAD results demonstrate a similar decrease-then-increase trend, with the optimal setting at 4 heads.
These provide guidance for setting the hyper-parameters of FilDeep in practice.
Moreover, we can note that the performance of FilDeep does not vary drastically under different hyper-parameter settings, showing its trustworthy robustness and stability.

\subsection{Visualization Analyses (RQ3)}

In this subsection, we aim to demonstrate the ability of our FilDeep to capture physical features and interaction patterns through visualization.
Specifically, our visualization analysis included two views: the cross-section reconstruction and the attention maps. 
They are used to intuitively showcase FilDeep’s performance in input feature extraction and physical interaction pattern extraction.

First, we analyzed the effective feature extraction of our encoders by using the reconstruction of cross-sections as an example.
As shown in Figure~\ref{fig:cross-section}, when faced with a new instance during testing, the CSE can effectively reconstruct its cross-section with minimal loss, demonstrating strong generalization ability in input feature extraction.
The strong capability of FilDeep’s encoders and decoders provides strong support for the overall performance.

\begin{figure*}[t]
    \centering
    \includegraphics[width=1.0\linewidth]{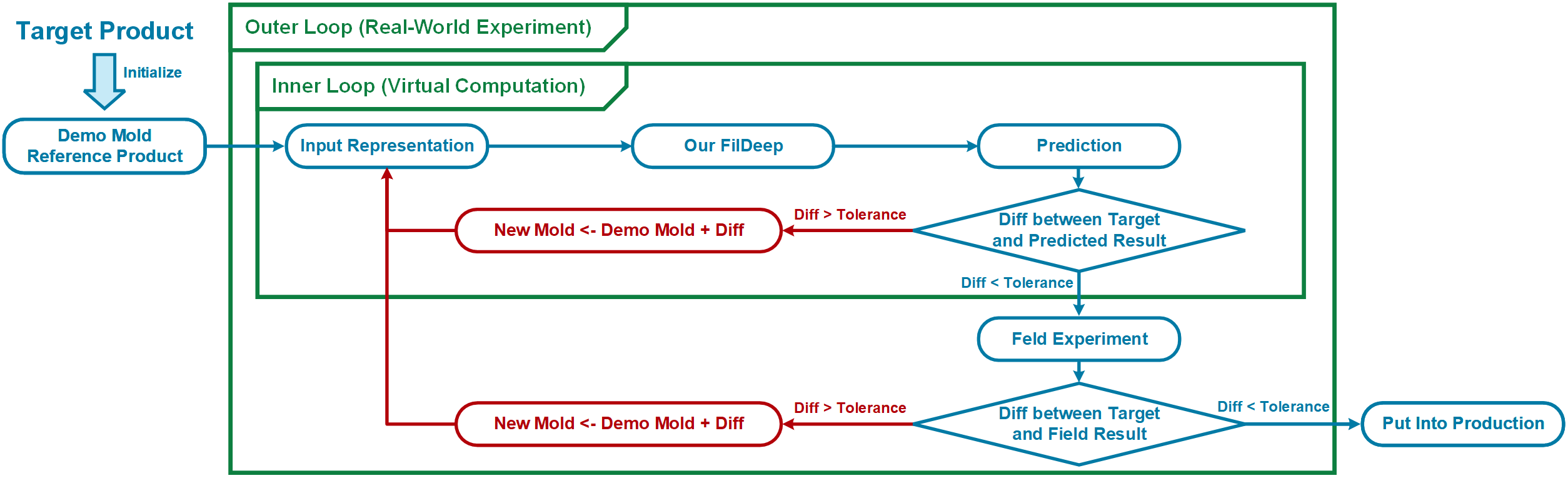}
    \caption{Detailed design of the Dual-Loop approach for practical applications.}
    \label{fig:Dual-Loop approach-new}
    \vspace{-2mm}
\end{figure*}

Second, we focused on the attention map within our ACF module, as shown in Figure~\ref{fig:attention}.
In the cross-attention map, the top-left and bottom-right corners (respectively corresponding to the head and tail of the workpiece and mold, where they make contact) are assigned higher weights.
This indicates that the contact relationships in the physical world are captured.
Moreover, in the self-attention map, more attention is given to the tail of the workpiece.
This aligns with domain knowledge, as the discrepancy between different fidelity accumulates at the tail of the workpiece.
As a consequence, this focus on the tail essentially captures the underlying relationship in the MF data.
These vivid observations indicate that, even without explicit constraints for training, our FilDeep can still learn the general interaction patterns from MF data.

\subsection{Real-world Deployment (RQ4)}

Collaborated with Fuyao Group, we have deployed FilDeep in a real manufacturing factory specializing in aluminum fabrications.
Previously, FEM~\cite{FEM} had been state-of-the-art in factories for years, suffering from poor efficiency.
A practical workpiece typically requires hours or even days for a single prediction, and the entire design process lasts 2 to 3 months, which is unacceptable in practice and highlights the necessity of our FilDeep.

Here, we compare the inference accuracy and efficiency of FilDeep with FEM in Figures~\ref{fig:acc-eff-1}-\ref{fig:acc-eff-2}. 
The results show that FEM, constrained by its accuracy-efficiency trade-off, exhibits a highly linear relationship on a logarithmic scale, as shown in Figure~\ref{fig:acc-eff-2}.
In contrast, our FilDeep represents a new paradigm, achieving up to $10^5$ faster inference efficiency than FEM while maintaining comparable accuracy, thereby fundamentally breaking the traditional efficiency-accuracy trade-off.
Meanwhile, we can also notice that FEM costs much more time for workpieces with more complex geometries.
Conversely, our FilDeep maintains consistently high inference efficiency regardless of the complexity of the geometry.
FilDeep's stable efficiency is also a very valuable property for practical applications.

In deploying FilDeep into practical applications, we developed a Dual-Loop approach, as illustrated in Figure~\ref{fig:Dual-Loop approach-new}.
It comprises the following two loops: the Inner Loop and the Outer Loop.
The Inner Loop iteratively refines the mold design and motion parameters through computational adjustments using FilDeep, whereas the Outer Loop, which is built upon the well-initialized design obtained from the Inner Loop, compensates for application errors through a few manual on-site adjustments.
More details are shown as follows.

\begin{itemize}
    \item \textbf{Inner Loop: Computation and Compensation.}
    Initially, based on the target product, we can initialize the demo mold by directly designing it to match the desired shape of the workpiece, and the corresponding motion parameters are uniquely determined using the classical involute method~\cite{curve}.
    Yet, due to the coexistence of elastic and plastic deformation, complex springback inevitably causes the final shape to deviate from the desired target.
    Thus, it is necessary to further adjust both the mold design and the motion parameters using the displacement compensation method.
    This Inner Loop continues until the error between FilDeep’s predicted result and the target falls below a predefined threshold.
    \item \textbf{Outer Loop: Deployment and Fine-Tuning.}
    Due to the inherent complexity and uncertainty of the manufacturing process, the gaps between computation and production (i.e., the sim-to-real discrepancies) are unavoidable.
    They can stem from various sources, such as: (a) Tolerances introduced by mold production machinery;
    (b) Installation errors during mold setup;
    (c) Zero-drift in loading arm motion parameters;
    (d) Temporal changes in workpiece material properties.
    These errors typically interact in complex ways, sometimes compounding or offsetting each other.
    In practice, the design provided by FilDeep's Inner Loop can effectively control the application error within about 10 mm.
    Then, on-site adjustments can be made based on the observed application error, including machining the mold using specialized tools and fine-tuning the loading arm parameters controlled by automated systems.
    These adjustments ensure that the final error is within an acceptable range for production.
\end{itemize}

\vspace{0.1cm}

Even considering the inevitable sim-to-real errors, the results determined by FilDeep can be controlled within approximately 6 mm to 10 mm during on-site testing, which is acceptable to be compensated for by on-site adjustments.
Through empirical production techniques, such as machining the mold with tools and fine-tuning the parameters of the loading arms, the final production error can be easily reduced to within an acceptable range.
Figure~\ref{fig:appendix-mold-new} shows a real mold for on-site production.
By deploying FilDeep, the entire design process is significantly shortened to about 1 week, achieving an 8- to 12-fold improvement in design efficiency.

\newpage

\section{Related Work}
\label{related}

MF learning presents an effective paradigm to learn from MF data, using LF predictions as conditioning inputs for HF refinement~\cite{MF-review, MF1, MF2}. 
While there are many early works applying MF learning to various applications, most of these efforts have been limited to directly applying simple linear models, e.g., Gaussian Process Regression (also known as Co-Kriging)~\cite{MF-review, MF-GP0, MF-GP1}.
Subsequently, some studies extended MF learning to neural networks, but they only address very limited scenarios: (1) structured low-dimensional regression using a multilayer perceptron that lacks capacity for geometric complexity~\cite{MF-PINN, ResMF, MF-active}, and (2) operator learning frameworks that focus on point-wise function-to-function mappings while neglecting critical physical interactions~\cite{MF-DeepONet0, MF-DeepONet1}.
These approaches are far from adequate for addressing our practical problems, which require tailored designs for representation and network modules.

\section{Conclusion and Discussion}
\label{conclution}
In this paper, we revealed a pervasive quantity-accuracy dilemma for DL in industrial applications, through the lens of a practical large deformation problem.
To resolve this challenge, we introduced FilDeep, the first DL framework to leverage MF data for learning large deformations of elastic-plastic solids.
Beyond addressing our immediate problem, FilDeep establishes an inspiring framework for resolving the quantity-accuracy dilemma, with promising potential to extend to a broader range of applications, such as electrode-electrolyte stability predictions, protein-ligand binding predictions, etc.
Moreover, as part of this work, we also contribute the first-ever MF dataset for large deformation problems to the community.
We believe this pioneering dataset can also effectively facilitate further significant research, e.g., cost-aware MF active learning.


\section*{Acknowledgments}
This work is supported partly by the National Natural Science Foundation of China (NSFC) under grant 62576013, the National Key Research and Development of China under grant 2024YFC2607404, the Jiangsu Provincial Key Research and Development Program under Grants BE2022065-1, BE2022065-3, and the Ningxia Domain-Specific Large Model Health Industry R\&D 2024JBGS001.
We also thank Yajiang Huang and Kejia Fan for their great contributions.

\normalem
\bibliographystyle{ACM-Reference-Format}
\bibliography{sample-base}

\clearpage

\appendix

\section{Appendix for Large Deformations of Elastic-Plastic Solids}
\label{appendix-deformation}
In the realm of continuum mechanics, deformation is defined as the change in the shape or size of a solid object, commonly referred to as strain~\cite{deformation}. 
This phenomenon is typically induced by external forces, known as load or stress, and is constrained by geometric boundaries~\cite{bend2}.
When deformations are extremely small, we can assume that the displacements of the material particles are much smaller (indeed, infinitesimally smaller) than any relevant dimension of the body.
This assumption forms the basis of the infinitesimal strain theory~\cite{theory}, where the material behavior can be effectively described using simple linear elastic models, such as Hooke's law.
When the strains become large enough that the assumptions of infinitesimal strain theory are no longer valid, such deformations are referred to as the large deformations~\cite{theory3}.
Evidently, due to their inherently complex nonlinear relationships, large deformation problems are more challenging than those involving small strains, making them the primary focus of this study.

\begin{figure}[t]
    \centering
    \includegraphics[width=1\linewidth]{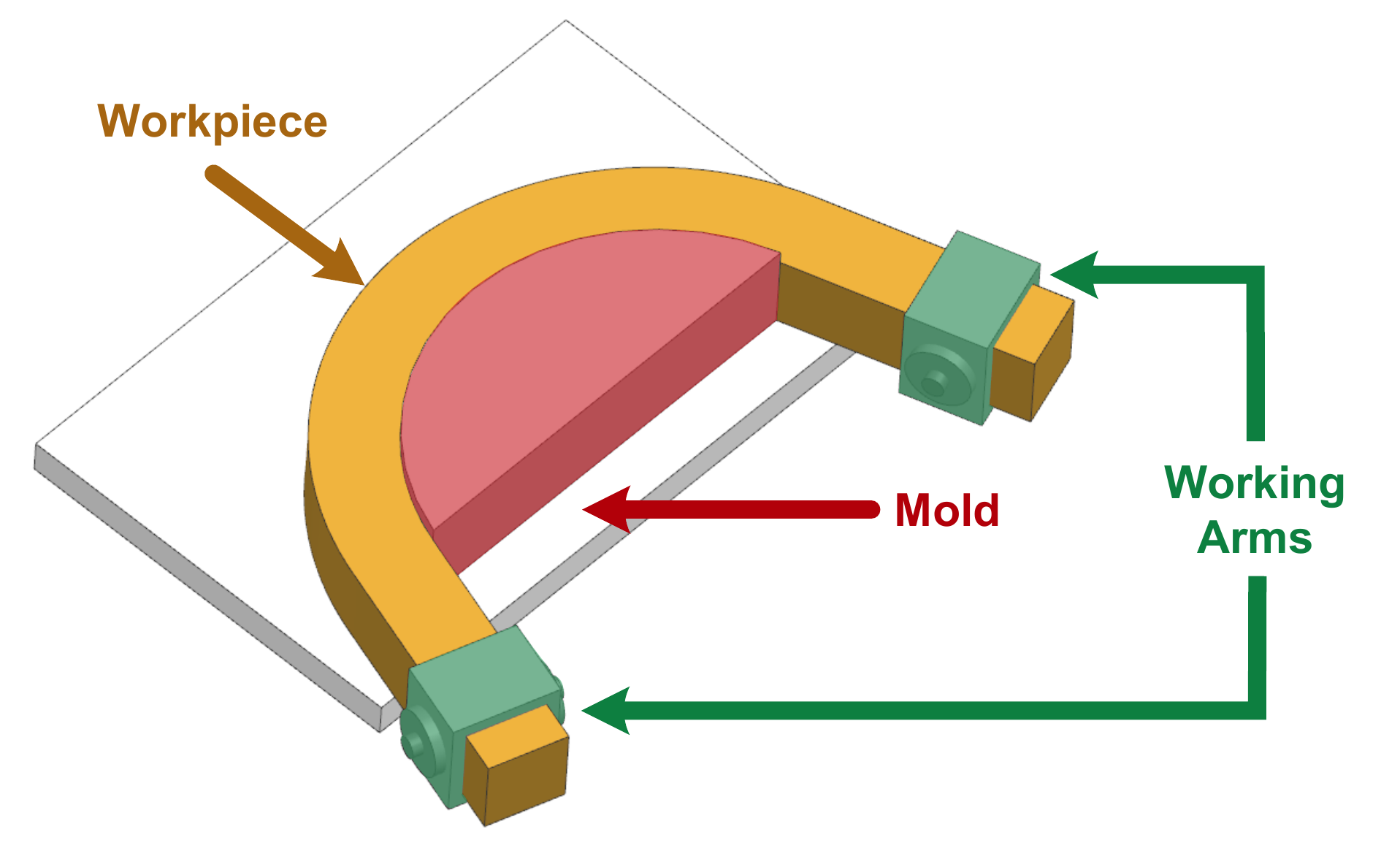}
    \caption{The stretch bending of metals.}
    \label{fig:appendix-bend}
\end{figure}

Large deformations can be either intentional or unintentional.
In this paper, we focus on stretch bending, one of the most popular metal fabrication techniques involving intentional large deformations~\cite{bending}.
As shown in Figure~\ref{fig:appendix-bend}, the stretch bending technique is extensively applied and holds significance in manufacturing applications.
In this process, the metal material (called the workpiece) is situated on a working machine and securely clamped by two loading arms.
When the working arms move and rotate, loads are applied to the workpiece, causing it to undergo large deformations to achieve the desired shape.
In this practical stretch-bending scenario, external forces are primarily applied to the workpiece through the loading arms, while geometric constraints are predominantly dictated by a pre-designed mold.

\begin{figure}[t]
    \centering
    \includegraphics[width=1\linewidth]{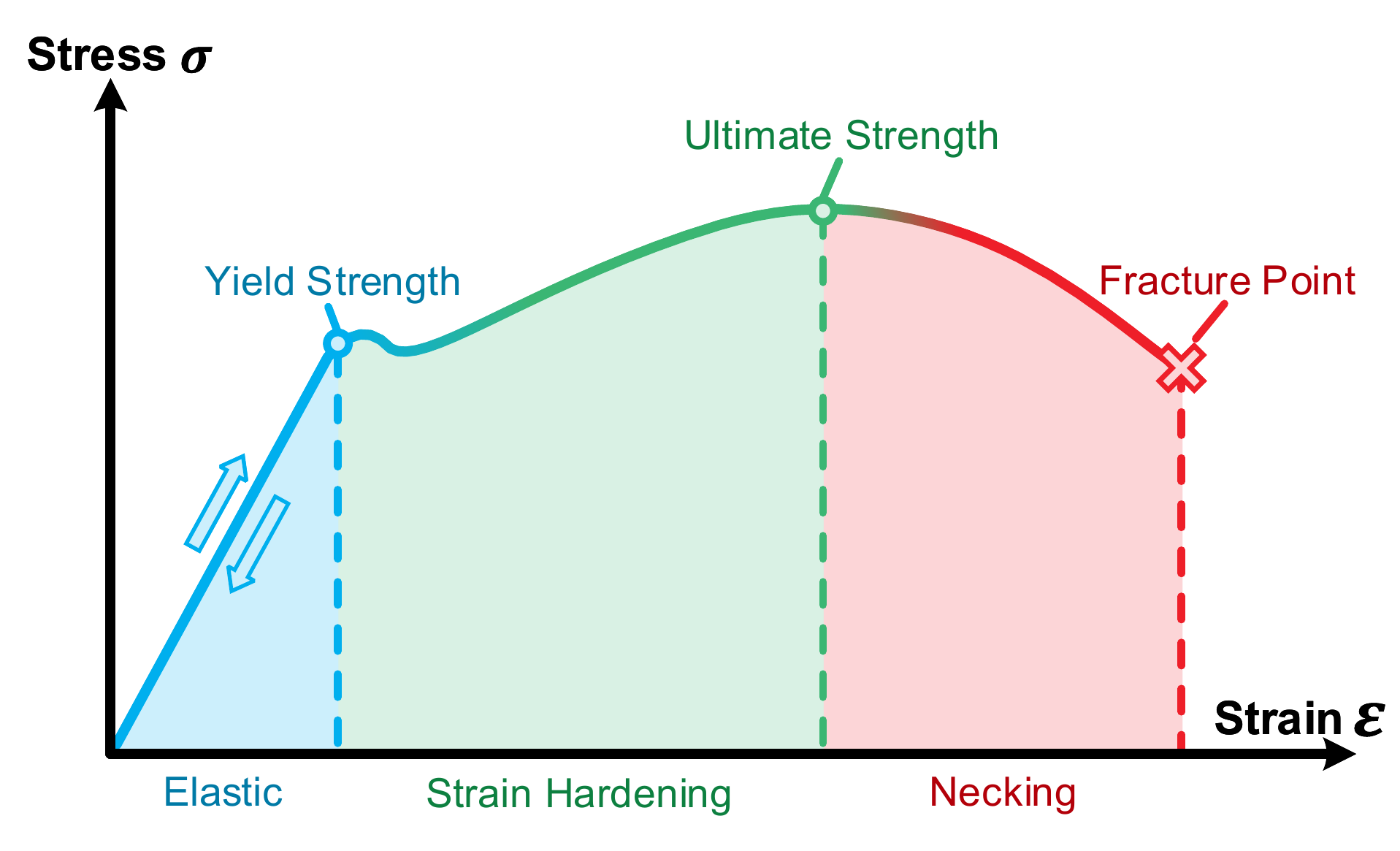}
    \caption{A stress-strain curve for a low-carbon steel.}
    \label{fig:appendix-stress-strain}
    \vspace{-0.2cm} 
\end{figure}

Given that the workpieces in stretch bending are metal solids (e.g., aluminum strips), the deformation process usually encompasses both elastic and plastic phases.
Figure~\ref{fig:appendix-stress-strain} illustrates the stress-strain curve for typical low-carbon steel, providing a visual representation of its elastic-plastic characteristics.
Specifically, in the case of elastic deformation, the material adheres to Hooke's law and can return to its original configuration upon the removal of the applied loads.
Conversely, if the applied loads are sufficiently significant to exceed the material's yield point, the material subsequently exhibits plastic behavior, resulting in permanent deformation.
It is important to note that after deformation, materials will experience a certain degree of springback upon the release of loads, but they will not fully revert to their original configuration due to the occurrence of plastic deformation.
The springback inherent in elastic-plastic materials necessitates careful design of the motion of molds and work arms to ensure that the workpiece deforms as intended.
In this context, the accurate prediction of such complex large deformations becomes crucial.

The metal stretch bending described above is typically regarded as a dynamical system that can be fundamentally characterized by a set of PDEs, governed by the constitutive laws of the specific metal~\cite{theory3}.
This perspective reflects a deterministic vision: the final steady state after deformation can, in theory, be predicted a priori if the material properties, initial configuration, and external conditions (e.g., applied loads and boundary constraints) are fully specified.
Unfortunately, due to the unknown material constitutive equations, irregular solution domains, and unpredictable multi-object contacts, the explicit forms of these PDEs are nearly impossible to write~\cite{theory2}.
Despite dedicated efforts by generations of physicists and engineers over more than half a century, a universally satisfactory solution remained elusive.


\vspace{3mm}
\section{Appendix for Traditional Numerical Methods}
\label{appendix-FEM0}
Traditional numerical methods for large deformations are mainly based on discretization techniques, such as Finite Element Method (FEM)~\cite{FEM}, Finite Difference Method (FDM)~\cite{FDM}, and Finite Volume Method (FVM)~\cite{FVM}.
Here, let's use the most popular method, FEM, as an example.
The original material will be partitioned into many small units (called elements), which is often referred to as the meshing technique~\cite{FEM}.
The large deformation process is considered in a step-by-step manner. In each step, the time unit and deformations are so small that the infinitesimal strain theory is valid~\cite{theory}, i.e., the geometry and the constitutive properties of the materials at each point are assumed unchanged.
With these assumptions, the constitutive equations of the solid are considerably simplified and approximated by Hook’s law, which is linear and solvable~\cite{theory}.

\begin{figure*}[t]
\setlength{\abovecaptionskip}{0.10cm}   %
    \centering
    \includegraphics[width=1.0\textwidth]{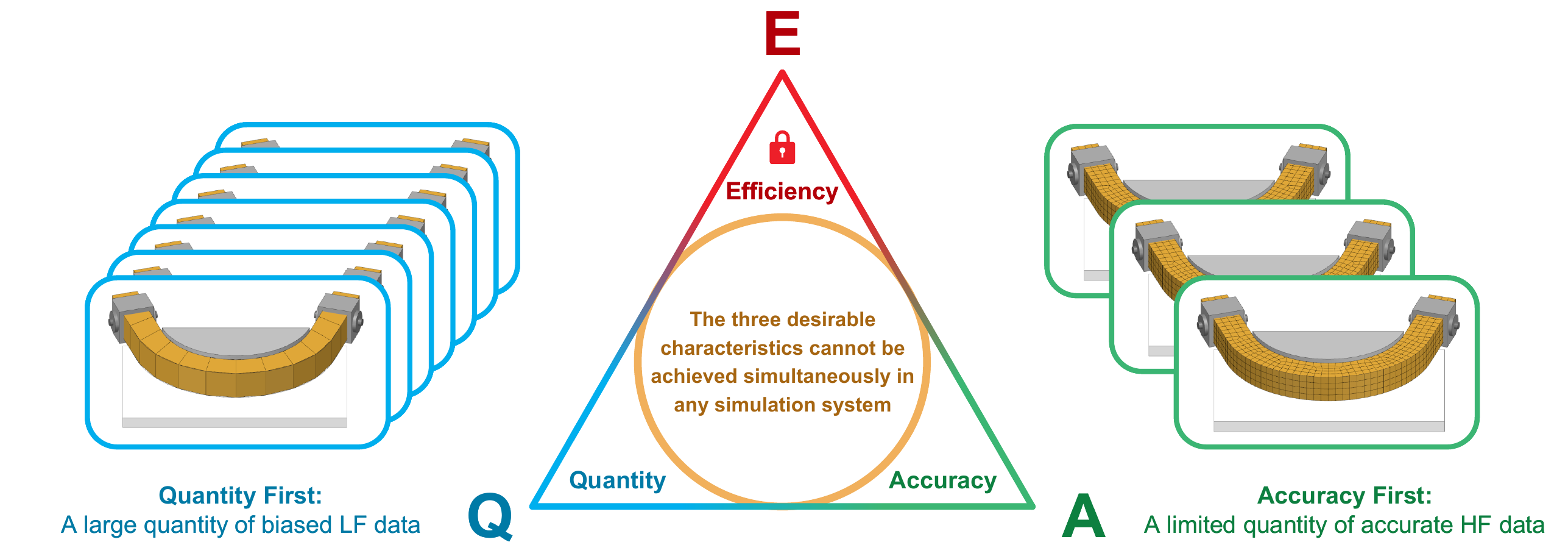}
    \caption{The quantity-accuracy dilemma within the EQA principle.}
    \label{fig:EQA}
    \vspace{-2mm}
\end{figure*}

With these measures, the effectiveness and efficiency of FEM are highly dependent on the appropriate meshing technique and sufficient incremental steps.
Specifically, the simulation accuracy is influenced by the regularity of the elements~\cite{Abaqus}.
High accuracy necessitates regular element shapes, such as regular tetrahedra or hexahedra, which, when dealing with raw materials of irregular shapes (e.g., a long but thin pillar), results in a large number of elements.
As the computations are with respect to each element, finer meshing with more elements and smaller time increments generally yields higher accuracy, and vice versa.
However, the enhanced accuracy necessitates a prohibitive volume of computations, creating an inevitable trade-off between accuracy and efficiency.
Simultaneously, since the entire process is conducted in a step-by-step manner that must be carried out sequentially, the inefficiency of traditional simulations cannot even be fully alleviated by parallel computing.
computations need to be carried out sequentially, which cannot be fully mitigated by parallel computing.
Notice that every problem instance has its unique input and output, and thus the computations must be restarted from scratch for every new instance.
This specificity further exacerbates efficiency concerns in practical applications.

\section{Appendix for Data Bottleneck with Quantity-Accuracy Dilemma}
\label{appendix-EQA}

As a kind reminder, our identified EQA principle states that {high efficiency, large quantity, and precise accuracy, as three desirable characteristics for industrial dataset generation, cannot be achieved simultaneously in any simulation system}.
Our EQA principle is, to some extent, very similar to the well-known CAP principle (i.e., Consistency, Availability, Partition tolerance) for distributed databases.
In fact, it empirically reveals the triangular trade-off relationship in the data construction process, as shown in Figure~\ref{fig:EQA}.

Specifically, when one of the three key factors is constrained, the remaining two form a trade-off.
\begin{itemize}
    \item In the first case, when the quantity is fixed, improving the accuracy of simulation data requires greater computational costs.
    Conversely, simplifying mesh or other simulation settings to reduce computational costs inevitably compromises simulation accuracy.
    This represents an efficiency-accuracy trade-off, which interestingly corresponds to one major limitation of traditional numerical methods mentioned earlier.
    During each inference with traditional numerical methods, the quantity can be considered fixed at a single instance. 
    Thus, its efficiency-accuracy trade-off is, in fact, a special case of our EQA principle.
    \item In the second case, when accuracy is fixed, reducing costs without sacrificing accuracy can only be achieved by limiting the data quantity.
    Conversely, increasing the data quantity without sacrificing accuracy leads to higher costs.
    This forms an intuitive efficiency-quantity trade-off, which is common in batch inference processes.
    For example, when using FEM to predict the deformation of similar types of materials under different external conditions multiple times, this trade-off becomes evident.
    A larger quantity of predictions results in higher costs, as the cost per instance remains roughly constant when accuracy is fixed.
    \item The third case is our primary focus, involving the quantity-accuracy trade-off when efficiency is constrained, as shown in Figure~\ref{fig:EQA}.
    Specifically, achieving high data accuracy requires costly HF simulations, resulting in low data quantity.
    Conversely, simulation acceleration by LF simulation simplifications can be used for high data quantity, but it comes at the expense of data accuracy.
    Thus, one can obtain either abundant biased LF data or a limited amount of accurate HF data for DL training.
    As a result, training deep learning models with single-fidelity data is inherently suboptimal, as it suffers from either high bias or high variance~\cite{bias–variance}.
\end{itemize}

Based on our previous analysis, existing DL methods often achieve suboptimal performance, as they fail to account for the quantity-accuracy trade-off. Whether prioritizing quantity or accuracy, these methods typically rely solely on single-fidelity data.
Therefore, in this paper, we introduce an interesting idea to simultaneously tackle both quantity and accuracy, with both LF and HF data.
By utilizing MF joint learning, rather than relying on a single fidelity level, our FilDeep method creates an opportunity to leverage both the HF data's accuracy advantage and the LF data's quantity advantage.

\end{document}